\documentclass[10pt,twocolumn,letterpaper]{article}

\usepackage[utf8]{inputenc} % allow utf-8 input
\usepackage[T1]{fontenc}    % use 8-bit T1 fonts
\usepackage{microtype}
\usepackage{times}
\usepackage{graphicx}
\usepackage{amsmath,amssymb,mathbbol}
\usepackage{algorithmic}
\usepackage[linesnumbered,ruled,vlined]{algorithm2e}
\usepackage{acronym}
\usepackage{enumitem}
\usepackage{balance}
\usepackage{xspace}
\usepackage{setspace}
\usepackage[skip=3pt,font=small]{subcaption}
\usepackage[skip=3pt,font=small]{caption}
\usepackage[dvipsnames]{xcolor}
\usepackage{booktabs,tabularx,colortbl,multirow,array,makecell}
\usepackage{pifont}
\usepackage{xr-hyper}
\usepackage[pagebackref,breaklinks,citecolor=uclablue,colorlinks=true,linkcolor=blue,bookmarks=false]{hyperref}
\usepackage[capitalise]{cleveref}

\usepackage{siunitx} % for the degree symbol
\usepackage{soul} % for \st command (strike-through) in collaborative writing
\usepackage[square,numbers]{natbib}
\makeatletter
\DeclareRobustCommand\onedot{\futurelet\@let@token\@onedot}
\def\@onedot{\ifx\@let@token.\else.\null\fi\xspace}
\def\eg{\emph{e.g}\onedot} 

\def\ie{\emph{i.e}\onedot}

\def\etc{\emph{etc}\onedot}

\def\etal{\emph{et al}\onedot}
\makeatother

\makeatletter
\renewcommand{\paragraph}{%
  \@startsection{paragraph}{4}%
  {\z@}{0ex \@plus 0ex \@minus 0ex}{-1em}%
  {\hskip\parindent\normalfont\normalsize\bfseries}%
}
\makeatother

\graphicspath{{figures/}}

\crefname{algorithm}{Alg.}{Algs.}
\Crefname{algocf}{Algorithm}{Algorithms}
\crefname{section}{Sec.}{Secs.}
\Crefname{section}{Section}{Sections}
\crefname{table}{Tab.}{Tabs.}
\Crefname{table}{Table}{Tables}
\crefname{figure}{Fig.}{Fig.}
\Crefname{figure}{Figure}{Figure}

\definecolor{gblue}{HTML}{4285F4}
\definecolor{gred}{HTML}{DB4437}
\definecolor{ggreen}{HTML}{0F9D58}

\definecolor{mygray}{gray}{.92}
\definecolor{uclablue}{rgb}{0.15, 0.45, 0.68}

\newcommand{\benchmark}{\texttt{ARNOLD}\xspace}
\newcommand{\sota}{state-of-the-art\xspace}

\newcommand{\cmark}{\ding{51}}%
\newcommand{\xmark}{\ding{55}}%

\definecolor{object}{rgb}{0.92,0.96,1.0}
\definecolor{scene}{rgb}{0.94,0.98,0.98}
\definecolor{state}{rgb}{0.92,0.98,1.0}
\definecolor{anystate}{rgb}{0.96,0.98,0.96}

\definecolor{teasercontinuous}{HTML}{2176ff}
\definecolor{teaserreal}{HTML}{dd6200}
\definecolor{teaserstate}{HTML}{db3a34}
\definecolor{teaserscene}{HTML}{1aa260}
\definecolor{teaserobject}{HTML}{963484}

\definecolor{NewGreen}{rgb}{0.5, 0.5, 0.5}

\newcommand{\mytest}{\ifmmode \mathrm{Yes}\else No\fi.}
\newcommand{\emojiobject}
{
    \vcenter
    {
        \hbox{
            \includegraphics[width=0.023\textwidth]{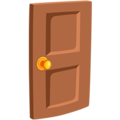}
        }
    }
}

\newcommand{\emojiscene}
{
    \vcenter
    {
        \hbox{
            \includegraphics[width=0.023\textwidth]{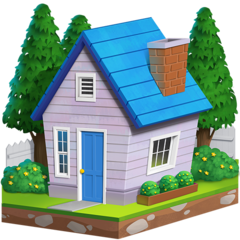}
        }
    }
}

\newcommand{\emojistate}
{
    \vcenter
    {
        \hbox{
            \includegraphics[width=0.024\textwidth]{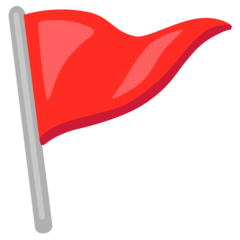}
        }
    }
}

\newcommand{\emojianystate}
{
    \vcenter
    {
        \hbox{
            \includegraphics[width=0.024\textwidth]{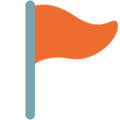}
        }
    }
}

\newcommand{\object}{\rowcolor{object}$\emojiobject \textit{Object}$}
\newcommand{\scene}{\rowcolor{scene}$\emojiscene \textit{Scene}$}
\newcommand{\state}{\rowcolor{state}$\emojistate \textit{State}$}
\newcommand{\anystate}{\rowcolor{anystate}$\emojianystate \textit{Any State}$}

\usepackage{iccv}
\usepackage{epsfig}
\usepackage[accsupp]{axessibility}  % Improves PDF readability for those with disabilities, recommended by camera-ready guidelines

\iccvfinalcopy % *** Uncomment this line for the final submission

\def\iccvPaperID{3397} % *** Enter the ICCV Paper ID here

\ificcvfinal\fi

\begin{document}

\title{\benchmark: A Benchmark for Language-Grounded Task Learning \\ With Continuous States in Realistic 3D Scenes}

\author{Ran Gong$^{1*}$,
Jiangyong Huang$^{2,5*}$,
Yizhou Zhao$^1$,
Haoran Geng$^{2,5}$,
Xiaofeng Gao$^1$,
Qingyang Wu$^4$, \\ \vspace{4pt}
Wensi Ai$^1$,
Ziheng Zhou$^1$,
Demetri Terzopoulos$^1$,
Song-Chun Zhu$^{2,3,5}$,
Baoxiong Jia$^5$,
Siyuan Huang$^5$ \\
$^1$University of California, Los Angeles,
$^2$Peking University,
$^3$Tsinghua University, \\
$^4$Columbia University,
$^5$National Key Laboratory of General Artificial Intelligence, BIGAI
}

\ificcvfinal\thispagestyle{empty}\fi

\twocolumn[{%we
\renewcommand\twocolumn[1][]{#1}%
\maketitle
\begin{center}
    \centering
    \vspace{-15pt}
    \captionsetup{type=figure}
    \includegraphics[width=\linewidth]{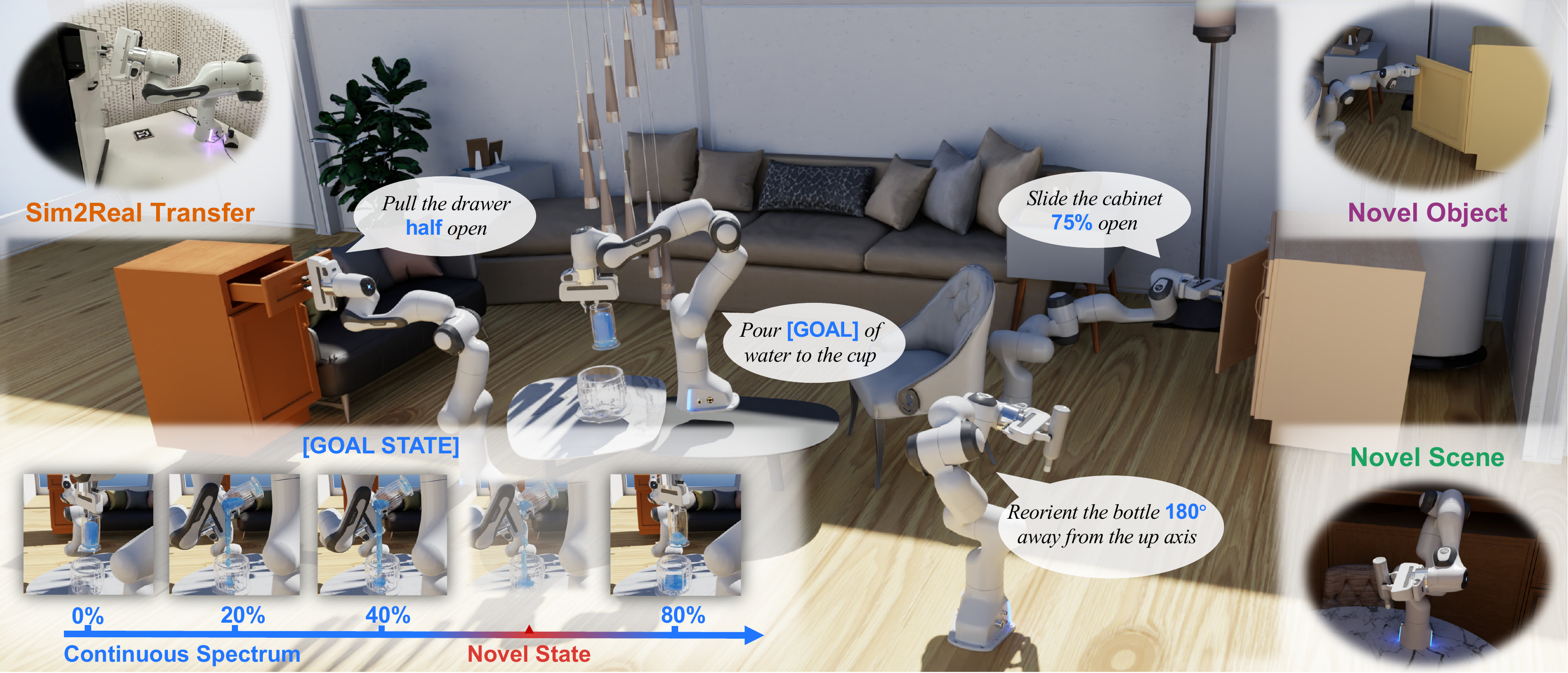}
    \captionof{figure}{\textbf{The \benchmark benchmark} for language-grounded task learning with \textcolor{teasercontinuous}{continuous states} in realistic 3D scenes. \benchmark provides 8 tasks with their demonstrations for learning and a testbed for the generalization abilities of agents over (1) \textcolor{teaserstate}{novel goal states}, (2) \textcolor{teaserobject}{novel objects}, and (3) \textcolor{teaserscene}{novel scenes}.
    % , as well as insights into (4) \textcolor{teaserreal}{Sim2Real transfer}.
    }
    \label{fig:overview}
\end{center}%
}]

\let\thefootnote\relax\footnotetext{* Equal contribution. Corresponding authors: Ran Gong, Jiangyong Huang, Baoxiong Jia, and Siyuan Huang.}

%%%%%%%%% 
\begin{abstract}
   Understanding the continuous states of objects is essential for task learning and planning in the real world. However, most existing task learning benchmarks assume discrete (\eg, binary) object goal states, which poses challenges for the learning of complex tasks and transferring learned policy from simulated environments to the real world. Furthermore, state discretization limits a robot's ability to follow human instructions based on the grounding of actions and states. To tackle these challenges, we present \benchmark, a benchmark that evaluates language-grounded task learning with continuous states in realistic 3D scenes. \benchmark is comprised of 8 language-conditioned tasks that involve understanding object states and learning policies for continuous goals. To promote language-instructed learning, we provide expert demonstrations with template-generated language descriptions. We assess task performance by utilizing the latest language-conditioned policy learning models. Our results indicate that current models for language-conditioned manipulations continue to experience significant challenges in novel goal-state generalizations, scene generalizations, and object generalizations. These findings highlight the need to develop new algorithms that address this gap and underscore the potential for further research in this area.\\ Project website: \href{https://arnold-benchmark.github.io}{https://arnold-benchmark.github.io}.
\end{abstract}

%%%%%%%%% BODY TEXT

\section{Introduction}

The ability to ground language is a crucial skill that has evolved over the course of human history, allowing people to learn and describe concepts, perform tasks, and communicate with one another. While recent developments have enabled the grounding of concepts in images~\cite{radford2021learning,kamath2021mdetr,saharia2022photorealistic, gao2016physical}, the interaction with physical environments \cite{deitke2020robothor, batra2020rearrangement, xia2020interactive, li2021igibson, nagarajan2020learning, jain2019two, jain2020cordial, gu2022multi, geng2022end, geng2023partmanip, xu2023unidexgrasp, wan2023unidexgrasp, deitke2022retrospectives, duan2022survey}, and language understanding in physical environments \cite{lu2019vilbert, anderson2018vision, pashevich2021episodic, zhang2021hierarchical,thomason2020vision, gao2022dialfred, wang2019reinforced, padmakumar2022teach}, few researchers have investigated the grounding of actions in daily tasks~\cite{shridhar2022cliport, zheng2022vlmbench, shridhar2022perceiver,wang2022humanise}. Given that humans can comprehend object status and relate language instructions to the physical environment, a pertinent question arises: \textit{How can we imbue robotic systems with the same capacity to understand and execute language instructions in the physical world?}

% One key capability that emerged from the long evolution of human beings is the ability to ground language. Such capability enables humans to describe and learn concepts, execute tasks, and also communicate with others. With the recent culmination of grounding concepts in images~\cite{radford2021learning,kamath2021mdetr,saharia2022photorealistic}, a few researchers have studied the grounding of actions~\cite{shridhar2022cliport, zheng2022vlmbench}. Considering how humans understand object status and relate language instructions to the physical world, a natural question to ask is: How can we give robotics systems the same capability to understand and execute language instructions in the physical world?

Learning action grounding in daily activities is a challenging task that presents several non-trivial difficulties. Firstly, robotic tasks rely heavily on detailed scene understanding for successful execution. This includes the understanding of geometry information, layouts, and visual appearances. The various combinations of scene configurations, including novel appearances, objects, and spatial positions further exacerbate this challenge. Therefore, it is crucial for robotic systems to acquire generalizable skills that can be transferred to different domains and settings.

% which can be further complicated by various combinations of scene configurations, including novel appearances, objects, and spatial positions~\cite{xing2021kitchenshift}. Therefore, the acquisition of generalizable skills that can be transferred to different domains and settings is critical for robotic systems.

% The learning of action grounding in daily tasks is a challenging task that presents several non-trivial difficulties. Firstly, robot tasks highly depend on detailed scene understanding for task execution, including the understanding of geometry information, layouts, and visual appearances~\cite{xing2021kitchenshift}.

Secondly, humans possess an exquisite ability to understand desired goal states precisely. This ability allows us to effortlessly map from simple descriptions (\eg, a cup half filled, a door fully opened, \etc) to the precise status of physical properties (\eg, half the volume, pulled to 180$^\circ$, \etc). However, it is exceedingly challenging for robots to learn the precise goal state from abstracted language instructions, especially when referring to an implicit range of continuous object states (\eg, a bit of coffee, slightly open, \etc) \cite{krantz2020beyond, hong2022bridging, mees2022matters}.  As a result, there is an urgent need for robot systems to maintain a mapping from language instructions to precise goal states in a continuous world.

A necessary first step toward tackling these challenges is to develop realistic robot simulation systems that enable language-grounded learning. Indeed, notable recent advances in simulated environments have facilitated grounded task learning~\cite{das2018embodied, shridhar2020alfred, mees2021calvin, zheng2022vlmbench, ma2022sqa3d}. Despite the impressive progress, these benchmarks suffer from several limitations that hinder the effective operation of robots in the real world: (1) They typically assume that tasks are performed in simple and clean environments, rather than in scenes that are spatially occupied by clutter and visually disturbed by diverse textured backgrounds~\cite{kumar2015mujoco, lin2020softgym, zheng2022vlmbench, shridhar2022perceiver}. (2) They assume discrete (\eg, binary) object states and perfect motor control that ignore the low-level geometry and dynamics of objects~\cite{szot2021habitat, srivastava2022behavior, ehsani2021manipulathor} and, consequently, they do not attempt in-depth physical state understanding or fine-grained manipulation skills. (3) These benchmarks do not ground instructions to precise states~\cite{zheng2022vlmbench, shridhar2022cliport}, thus neglecting the challenging problem of grounding language to object states.
To address these critical challenges of language-grounded robot task learning, we introduce a new benchmark, \benchmark, for grounding task instructions to precise robot actions and object states in realistic natural scenes (\cref{fig:overview}). Specifically, we leverage a highly accurate physics simulation engine to create eight challenging robot manipulation tasks that include continuous robot motion, friction-based grasping, and a variety of object state manipulations. Each task is associated with a set of goals sampled from a continuous range of object states and their corresponding detailed task descriptions in human language form. We further provide plentiful demonstrations of each task with trajectories generated with a template-based planner for robot learning. 
 
% About the tasks, evaluation
% allows for the grounding of task instructions to \textbf{continuous} robot actions and object states in an environment that closely resembles reality \cite{li2022behavior}. 
% This is made possible by the use of the powerful Nvidia Issac Sim platform, which enables highly accurate physics simulations (including fluids and soft body simulations) and state-of-the-art visual rendering.

% To synthesize realistic demonstrations for eight distinct tasks within the \benchmark benchmark, we have developed a hybrid template-based approach. These tasks involve continuous robot motion and friction-based grasping, as well as a variety of object state manipulations, requiring a range of different motor skills, such as grasping, pushing, pulling, and pouring liquids into containers.

% To ground the task instructions to the robot actions and object states, we have paired each demonstration with a template-based language instruction that clearly and concisely describes the task goals, allowing the model to learn from the instruction and the corresponding demonstration to build its knowledge and understanding of the task at hand.

% \bx{are we going to describe the generalization splits? (the following argument) we might also need to rephrase the contribution section to make the contribution of the workshop submission version more clear.}

% \sy{Expand: how to evaluate generalizations, can put the 40 objects and 20 scenes here}
To provide an in-depth evaluation of language-grounded task learning, we complement prior research with an evaluation that targets the ability of agents to generalize learned language-grounded skills to unseen scenarios, including novel scenes, novel objects, and our featured novel goal states. We have meticulously curated a collection of 40 distinctive objects and 20 scenes from open-source datasets~\cite{xiang2020sapien,kolve2017ai2,fu20213d} and designed data splits for evaluating different aspects of agents' generalization ability in language-grounded task learning. Furthermore, we provide thorough experimental analyses and show that \sota language-conditioned manipulation models still suffer with regard both to grounding and generalization. Additionally, we show that state modeling is crucial for tasks in \benchmark through carefully designed ablation studies.
% \jy{the following needs rephrase}studying the ability of models to generalize to arbitrary goal states within a continuous range.  

% In contrast to prior research, \benchmark evaluates the capacity of an agent to generalize to previously unseen object states, in addition to novel objects, by leveraging its understanding of language instructions and continuous object states. For this purpose, we meticulously curated a collection of 40 distinctive objects and 20 scenes sourced from open-source datasets such as \cite{xiang2020sapien, kolve2017ai2, fu20213d}. Our team obtained demonstrations through a combination of human-annotated objects, robot positions, and motion planning. Apart from the standard train/validation/test split, we further separated out five scenes as novel scene configurations, two objects as novel objects, and one object state per task category as a novel state, for testing the model's performance on these holdout sets.

% \sy{Experiments and conclusions, what are the hints for future research}
% According to the results of our experimental analysis, current language-conditioned manipulation models exhibit persistent challenges when attempting to generalize across various splits. Additionally, we have discovered that incorporating continuous state prediction as a component of the loss function can enhance model performance, thereby indicating that a greater understanding of continuous state representation has the potential to improve model performance.

In summary, \benchmark makes the following contributions: 
\begin{itemize}[leftmargin=*,noitemsep,nolistsep]
\item A \textbf{realistic 3D interactive environment} with diverse scenes, objects, and \textbf{continuous object states}, facilitating the learning and evaluation of precise robot manipulation.
\item A systematic benchmark comprising eight challenging \textbf{language-grounded robotic tasks} and evaluation splits for different aspects of \textbf{skill generalization}.
% for learning language-grounded and generalizable manipulation policies.
\item \textbf{Extensive experiments and analyses} of \sota language-conditioned manipulation models, revealing their strengths and weaknesses in promoting future research on language-grounded task learning. 
% We further demonstrate the potential of zero-shot \textbf{Sim2Real} transfer when training on \benchmark. \jy{the potential of Sim2Real transfer}.
\end{itemize}

% \begin{itemize}[leftmargin=*,noitemsep,nolistsep]
%     \item A set of eight different \textbf{language-conditioned manipulation} tasks featuring different motor skills and diverse object states. 
%     \item A \textbf{realistic} 3D interactive environment with \textbf{continuous states} for different objects and fluids with \textbf{human-annotated} robot locations and interactive object locations.
%     \item A systematic benchmark for \textbf{language} generalization, \textbf{object} generalization, and \textbf{scene} generalization for language conditioned manipulation policies.
%     \item An extensive \textbf{analysis} of the current state-of-the-art language conditioned manipulation models.% We hypothesize that the task difficulty may be attributed to the models' inability to ground textual descriptions to continuous object states. 
%     \item \sy{benchmarks and results.}
% \end{itemize}

% \begin{figure}[h]
%      \centering
%      \begin{subfigure}[b]{0.49\textwidth}
%          \centering
%          \includegraphics[width=\textwidth]{figures/fluid4.png}
%      \end{subfigure}
%      \hfill
%      \begin{subfigure}[b]{0.49\textwidth}
%          \centering
%          \includegraphics[width=\textwidth]{figures/teaser.png}
%      \end{subfigure}
%      \hfill
%     \caption{Teaser \sy{For water-related tasks, can we show them on a drawer or table rather than the ground?}}
%     \label{fig:teaser}
% \end{figure}

%-------------------------------------------------------------------------
\section{Related Work}

\begin{table*}[t!]
\centering
\caption{\textbf{Comparison with existing benchmarks.} \benchmark features language-grounded robot control over continuous object states with a large number of demonstrations in photo-realistic scenes. \benchmark also \textbf{leverages advanced physics simulations powered by PhysX 5.0 to simulate articulated bodies and fluids}. \textbf{Language:} Task goals are specified by human language instruction.  \textbf{Multi-Camera: } Robot is equipped with multiple cameras. \textbf{Fluid:} Advanced fluid simulation. \textbf{Physics:} Realistic physics simulation with realistic grasping. $^1$:~RLbench-based benchmarks use simplified grasping.  \textbf{Continuous:} Object state and goal state are continuous. \textbf{Scene:} Tasks are performed with a realistic scene background. \textbf{Robot:} Perform actions with real robots for all tasks. \textbf{R}: Rasterization.  \textbf{RT}: RayTracing. \textbf{Flexible Material}: Easy to change materials and textures. \textbf{Generalization}: Systematic generalization test at different levels.}
\vspace{0.5em}
\small
\resizebox{\linewidth}{!}{
\begin{tabular}{lcccccccccc}
\toprule
Benchmark & Language & Multi Camera & Fluid & Physics & Continuous & Scenes & Robot & Rendering  & Flexible Material & Generalization \\
\midrule
Alfred~\cite{shridhar2020alfred} & \cmark & \xmark & \xmark & \xmark & \xmark & \cmark & \xmark & R & \xmark & \xmark \\
Maniskill~\cite{mu2021maniskill, gu2023maniskill2} & \xmark & \cmark & \cmark & \cmark & \cmark & \xmark & \cmark & R  & \xmark & \cmark \\
Calvin~\cite{mees2021calvin} & \cmark & \cmark & \xmark & \cmark & \xmark & \cmark & \cmark & R  & \xmark & \cmark \\
Behavior~\cite{li2022behavior, srivastava2022behavior} & \xmark & \xmark & \cmark & \cmark & \cmark & \cmark & \cmark & RT  & \cmark & \xmark \\
KitchenShift~\cite{xing2021kitchenshift} & \xmark & \xmark & \xmark & \cmark & \xmark & \xmark & \cmark & R & \xmark & \cmark \\
RLBench \cite{james2020rlbench} & \cmark & \cmark & \xmark & \cmark$^{1}$ & \xmark & \xmark & \cmark & R  & \xmark & \xmark \\
Softgym~\cite{lin2020softgym} & \xmark & \xmark & \cmark & \cmark & \cmark & \xmark & \xmark & R & \xmark & \xmark \\
Orbit \cite{mittal2023orbit} & \xmark & \xmark & \cmark & \cmark & \xmark & \cmark & \cmark & RT & \cmark & \xmark \\
Vlmbench~\cite{zheng2022vlmbench} & \cmark & \cmark & \xmark & \cmark$^{1}$ & \cmark & \xmark & \cmark & R & \xmark & \xmark \\
Ravens~\cite{zeng2021transporter, shridhar2022cliport} & \cmark & \cmark & \xmark & \cmark & \xmark & \xmark & \cmark & R & \xmark & \cmark \\
Habitat HAB \cite{szot2021habitat} & \xmark  & \xmark & \xmark & \xmark & \xmark & \cmark & \cmark & R & \xmark & \cmark \\
TDW Transport \cite{gan2020threedworld, gan2022threedworld} & \xmark & \xmark & \xmark & \xmark & \xmark & \cmark & \cmark & R & \xmark & \xmark \\
\midrule
\textbf{\benchmark} & \cmark & \cmark & \cmark & \cmark & \cmark & \cmark & \cmark & RT & \cmark & \cmark \\
\bottomrule
\end{tabular}
}
\label{tab:comparison}
\end{table*}

\paragraph{Simulators for Embodied AI.} Significant progress has recently been made in developing simulators for training and evaluating AI agents to perform indoor household activities \cite{li2021igibson, fu2022rfuniverse, gan2020threedworld, chen2022soundspaces, deitke2022procthor, chang2017matterport3d}. To mitigate complexity, most of these simulators make simplifications about world states and actions, abstracting robot manipulation into symbolic planning in discrete action \cite{puig2018virtualhome, kolve2017ai2} and state spaces. However, agents trained in such settings are unaware of the relationship between actions and the geometries and dynamics of objects, therefore limiting their abilities in real-world scenarios. Recent efforts have gradually transitioned to continuous action spaces, but they still make some simplifications. For example, grasping is often simplified by attaching a nearby object to the gripper \cite{ehsani2021manipulathor, srivastava2022behavior}, or through contact~\cite{li2022behavior, james2020rlbench, szot2021habitat}. 

Among works that provide continuous object state change simulation, most do not focus on manipulating object states in a precise and fine-grained manner. For example, VRKitchen \cite{gao2019vrkitchen} defines task goals in a discrete manner even though the underlying object states are continuous. Softgym~\cite{lin2020softgym} is an object manipulation benchmark that provides a realistic simulation of deformable objects; however, it lacks diversity among objects and scenes. 

% In addition, as Isaac Sim is a part of Omniverse, authoring more complex assets can be easily managed through other applications within the Omniverse universe as indicated by \cite{mittal2023orbit, li2022behavior}
% Most of these simulators only simulate preconditions and post-effects of agent actions or use simplified state representations. Instead, A large number of benchmarks use simplified abstract discrete action space. Abstract discrete action space can reduce the task's difficulty. 
 % This limits agents' capability to transfer learned skills to real-world scenarios. For example, grasping is often simplified by attaching a nearby object to the gripper \cite{ehsani2021manipulathor, shridhar2020alfred, srivastava2022behavior, szot2021habitat}, or through contact \cite{li2022behavior, james2020rlbench}. 
By contrast, \benchmark provides a wide variety of scenes and objects. And \benchmark simulates continuous states for articulated objects and simulates fluids at the particle level. We control the robots with 7-DOF continuous control and friction-based grasping powered by a state-of-the-art physics engine (PhysX 5.0). 
Whereas most of the environments \cite{gu2023maniskill2, szot2021habitat} optimize for speed, we optimize for the realism of the rendering. And \benchmark also equips a remarkable rendering speed at 185 FPS (37 FPS with five cameras).
% With Sim2Real experiments, we demonstrate that it is possible to transfer policies learned from our environments directly to \textbf{novel objects} in the real world.

% \citet{deitke2020robothor} suggests that even though they are realistic, simulation environments cannot be compared to the real world, as models trained in the real world do not have similar performance when tested in the virtual world. In this work, we leverage ray tracing and accurate continuous physics simulation to generate photo-realistic images and try to close the gap between sim to real.

\paragraph{Language Conditioned Manipulation.} Relating human language to robot actions has been of recent interest \cite{ lynch2020language, stepputtis2020language,zheng2022vlmbench, jiang2022vima, deng2020mqa, huang2022inner,huang2023diffusion, nair2022learning}. However, the environments in these efforts either lack realistic physics \cite{shridhar2020alfred, shridhar2022cliport} or do not have realistic scenes \cite{shridhar2022perceiver, mees2021calvin} where the surroundings of the agent will constrain its motion, and different scene objects might occlude the agent's viewpoint. Additionally, systems like \cite{huang2022inner, brohan2022can, driess2023palme, lynch2022interactive} are application-based, lacking a systematic benchmark for language-conditioned manipulation. Most importantly, prior work aims to ground human language to static object properties, such as colors and shapes. By contrast, \benchmark provides instructions for continuous object states. We compare between \benchmark and other related benchmarks in \cref{tab:comparison}.

\paragraph{Continuous State Understanding.} Some recent research tries to predict object states \cite{liu2017jointly, nagarajan2018attributes}. However, the object states are discrete rather than continuous. More recently, researchers \cite{weng2021captra,di2022gpv, wei2022self, tseng2022cla} tried to predict object states continuously. However, they do not address the manipulation of objects from arbitrary starting states to the desired states. Moreover, they do not model the language grounding process. Most recently, Ma \etal~\cite{ma2023sim2real2} propose a method to perform precise object state manipulations, but their approach does not perform language grounding and only a small subset of our tasks are covered with their simple motion primitives.

Compared with prior works, we provide more diverse goal states to cover the continuous state space instead of learning only binary goal states. This allows models to understand the continuous state space. On the other hand, we also propose evaluation of generalization in terms of continuous state understanding (see \cref{sec:benchmark}). This evaluates how the model leverages its understanding of the state space to generalize within a continuous spectrum, which is rarely studied in prior works. Though more goal states can be added with our continuous simulation, we leave \benchmark at the current scale since more states will lead to higher costs of data generation due to the compositions of object/scene/state.

\section{The \benchmark Benchmark}

The \benchmark benchmark is motivated by the abilities that an intelligent manipulator agent should possess, including (1) the ability to comprehend and ground human instructions to precise world states, (2) the capacity to acquire policies for generating accurate actions and plans toward precisely defined goal states, and (3) the feasibility of transferring such abilities to the real world. Therefore, in \benchmark we focus on language-conditioned manipulation driven by continuous goal states situated in diverse photo-realistic and physically-realistic 3D scenes.

% In \cref{sec:simulator}, we will introduce the simulation environment, which is able to provide complex scenarios to approximate the real world yet poses more challenges for the learning. In \cref{sec:task}, we will introduce the formulation of eight tasks in \benchmark. After injecting continuous state, these tasks require the sense of grounded goal state as well as how states evolve, in addition to the conventional manipulation skills (\eg, grasp, rotate).

% The goal of \benchmark is to evaluate the learning of language-conditioned continuous control policy over a diverse range of physical-realistic and photo-realistic 3D scenes. In this setting, an agent needs to have the following capabilities: 

% \begin{itemize}[leftmargin=*, noitemsep, nolistsep]
%     \item Understand the goal state of the current object from human language instruction.
%     \item Estimate current object states from multiple camera inputs.
%     \item Propose and actuate motion plans based on physics over long-horizon.
% \end{itemize}

 % Concretely, for an instruction like pouring half of the water out of the cup, the agent needs to understand the instruction and know what the cup should look like when it's half full from multi-model sensors and reasoning. Then it needs to further reason over the best sequence of actions to grasp the cup and pour the water out. Finally, it needs to know when and how to stop to meet the goal object state it determined previously. 
 
\subsection{Simulation Environment}
\label{sec:simulator}

\begin{figure}[t!]
    \centering
    \includegraphics[width=\linewidth]{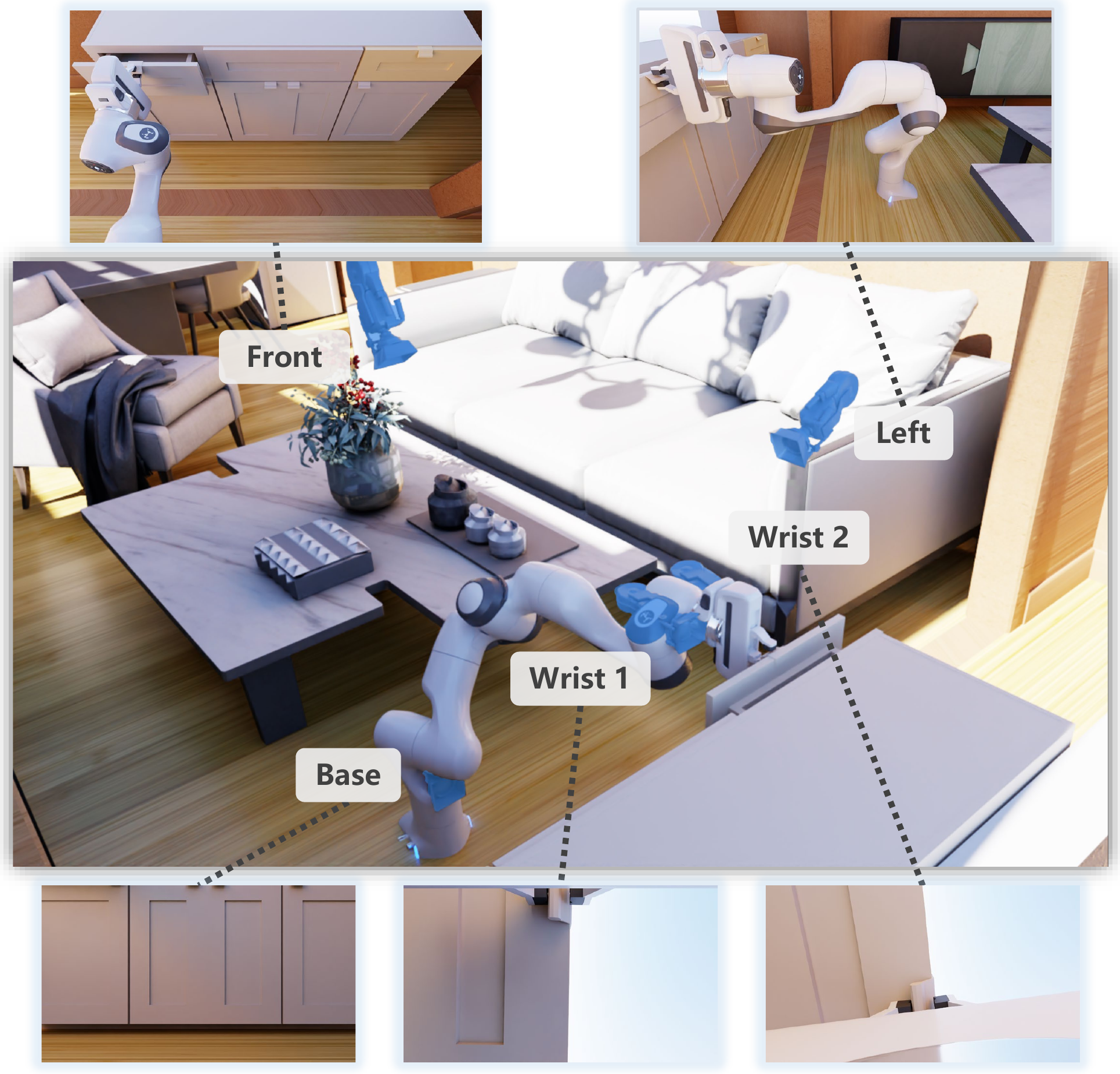}
    \caption{Multi-view robot observation in \benchmark. The top row shows views from the front and left cameras and the bottom row from the base and two wrist cameras.}
    \label{fig:multi-view}
\end{figure}

% \subsubsection{Environment Components}
\paragraph{Simulation Platform.} \benchmark is built on NVIDIA's Isaac Sim \cite{makoviychuk2021isaac}, a robotic simulation application that provides photo-realistic and physically-accurate simulations for robotics research and development. In \benchmark, the photo-realistic rendering is powered by GPU-enabled ray tracing, and the physics simulation is based on PhysX 5.0. \cref{fig:overview} and \cref{fig:multi-view} provide examples of simulation and rendering. 
% As demonstrated by \cite{li2022behavior}, the rendering effect provided by Isaac Sim is much more realistic than any existing simulator.

\paragraph{Physical Simulation.} To ensure physically-realistic simulation, we assign physics parameters to objects, including weight and friction for rigid-body objects, and cohesion, surface tension, and viscosity for fluids. These parameters are selected as in prior work~\cite{mu2021maniskill} and are adjusted by human operator feedback. Fluids are simulated using the GPU-accelerated position-based-dynamics (PBD) method \cite{macklin2013position} through NVIDIA's Omniverse platform. Depending on the rendering speed, we perform an optional surface construction process using marching cubes \cite{lorensen1987marching} to achieve the final fluid rendering effect.
% \textcolor{blue}{XF: describe how fluids are simulated. More details are needed in the supplementary.}

\paragraph{Scene Configuration.} There are 40 distinct objects and 20 diverse scenes in \benchmark. The scenes are curated from \cite{fu20213d}, a large-scale synthetic dataset of indoor scenes. This endows \benchmark with professionally designed layouts and high-quality 3D models. In addition to objects provided by Isaac Sim, we collected objects from open-source datasets \cite{kolve2017ai2,xiang2020sapien}. We modified object meshes to enhance visual realism, \eg, by modifying materials and adding top covers to cabinets and drawers. For more stable physics-based grasping, we performed convex decomposition to create precise collision proxies for each object. More details are found in Appendix~A of \cite{gong2023arnold}.

\paragraph{Robot.} We use a 7-DoF Franka Emika Panda manipulator with a parallel gripper in \benchmark for task execution. The agent has direct control over its seven joints and its gripper. We represent end-effector actions with three spatial coordinates for translation and quaternion for rotation, as it is more tractable for policy learning \cite{liu2022frame}. We utilize the built-in motion planner of Isaac Sim to transform the end-effector action back to the space of robot joints for execution. Currently, our tasks do not involve navigation, \ie, the robot base remains fixed during task execution.

\paragraph{Visual Input.} In \benchmark, we use five cameras around the robot for visual input. As shown in~\cref{fig:multi-view}, the cameras provide various views, including front, left, robot base, and wrist. While each camera provides RGB-D input at a resolution of $128\times 128$ by default, users can render at arbitrary resolution. Notably, unlike the deterministic rendering in prior works \cite{shridhar2022perceiver,zheng2022vlmbench}, the rendering in \benchmark is stochastic due to the ray tracing sampling process \cite{shirley2008realistic}, which makes \benchmark more realistic and challenging. In addition to the visual observation, other auxiliary observations can be accessed, \eg, camera parameters, robot base pose, and part-level semantic mask. Other Omniverse sensors (\eg, tactile) are excluded here since they are not required by the tasks and models. They are all available if necessary.

\subsection{Task Design}
\label{sec:task}

% \begin{figure}[ht]
%     \centering
%     \includegraphics[width=0.5\textwidth]{figures/ArnoldTeaser.png}
%     \caption{Examples of tasks in \benchmark. For accomplishing these tasks, it requires visual recognition, text  understanding, as well as diverse cognitive skills.}
%     \label{fig:task_vis}
% \end{figure}
% \sy{Figure out how to balance 3.2 and 3.4 here}
We include eight tasks with various goal state variations in \benchmark. Specifically, we focus on continuous goal states and define success ranges around them wherein robots should maintain object states for 2 seconds to succeed.
%evaluate executions by checking if the manipulation of a robot has successfully changed the object state to be within a success range around the goal state.
\cref{tab:task_type} provides an overview and \cref{fig:overview} a visualization. More illustrative examples are shown in Appendix~B of \cite{gong2023arnold} . Performing these tasks requires capabilities in language grounding, friction-based grasping, continuous state understanding, and robot motion planning. Additional task details follow:

\begin{table}[t!]
\centering
\caption{Overview of the 8 tasks in \benchmark. Each task features 4 goal states specified by human language, one of which is reserved for novel state evaluation. The task is deemed successful when the object state remains in the success range for two seconds. Note that \textsc{TransferWater} imposes the extra condition that only less than $10\%$ of the original amount of water in the cup can be spilled.}
\vspace{0.5em}
\small
\resizebox{\linewidth}{!}{
\begin{tabular}{llc}
\toprule
\textbf{Task Types} & \textbf{Goal States}  & \textbf{Success Ranges} \\
\midrule
\textsc{PickupObject} & $10,20,30,40$ (cm) & $\pm 5$\,cm \\
\textsc{ReorientObject} & $0,45,135,180$ (\textdegree) & $\pm20$\textdegree$\ \ $\\
\textsc{OpenDrawer} & $25,50,75,100$ ($\%$) & $\pm10\%$ \\
\textsc{CloseDrawer} & $0,25,50,75$ ($\%$) & $\pm10\%$ \\
\textsc{OpenCabinet} & $25,50,75,100$ ($\%$) & $\pm10\%$ \\
\textsc{CloseCabinet} & $0,25,50,75$ ($\%$) & $\pm10\%$ \\
\textsc{PourWater} & $25,50,75,100$ ($\%$) & $\pm10\%$ \\
\textsc{TransferWater} & $20,40,60,80$ ($\%$) & $\pm10\%$ \\
\bottomrule
\end{tabular}
}
\label{tab:task_type}
\end{table}

\begin{itemize}[leftmargin=*,noitemsep,nolistsep]
    \item In \textsc{PickupObject} and \textsc{ReorientObject}, we instruct the robot to manipulate a bottle to achieve different goals. For the former, the initial state of the object is on the ground with goals specifying heights above the ground. For the latter, the initial state of the object is on the ground, oriented horizontally (the state value equals 90\textdegree), with goals specifying the angle between the object's orientation and the upright orientation.
    \item In the four tasks \{\textsc{Open},\textsc{Close}\}\hspace{-2pt}\{\textsc{Drawer},\textsc{Cabinet}\}, the goal value specifies the geometric state of the articulated joint,  either in terms of distance (for prismatic joints in \textsc{Drawer}) or angle (for revolute joints in \textsc{Cabinet}). The initial state is any value smaller than the goal for \textsc{Open} and larger than the goal for \textsc{Close}.
    % \bx{random selected?}.
    \item In \textsc{PourWater} and \textsc{TransferWater}, the manipulated object is a cup containing water, and the goal specifies the percentage of water to be poured out (\textsc{Pour}) or poured into another cup (\textsc{Transfer}). In these two tasks, the goal values are specified as percentages of water relative to the initial amount of water in the cup.
\end{itemize}

% \bx{Need to mention goal variation here so that readers don't get lost in the discussion on evaluation split and need to balance where to discuss the succ. condition, if it's still going to be discussed in metrics, we might want to put tab.2 and 3 together side-to-side and make the caption of tab.2 shorter}
Our task pool covers a variety of manipulation skills and grounding aspects. \textsc{PickupObject} and \textsc{ReorientObject} are selected for the basic skills of moving and rotating objects and the grounding of distances and angles. These abilities are then composed and reinforced in the four tasks \{\textsc{Open},\textsc{Close}\}\hspace{-2pt}\{\textsc{Drawer},\textsc{Cabinet}\}, where the goal state is grounded on the state of the manipulated drawer or cabinet joint. Beyond rigid-body objects, fluid manipulation in the two tasks \{\textsc{Pour},\textsc{Transfer}\}\hspace{-2pt}\textsc{Water} challenges the robots' ability to manipulate containers and move fluid, grounding goal state values to fluid volumes.

\subsection{Data Collection}

% \paragraph{Human demonstrations.} To handle the aforementioned tasks, We collect \textbf{nearly 2k human demonstrations} for imitation learning. 
% % This approach is preferred to reinforcement learning due to sample efficiency. 
% In these demonstrations, there are rich variations for each task, \eg, objects, scenes, goal states, and positions. However, it is intractable to directly use these human demonstrations as the episodes are too dense to manifest high-level control policy. It is also costly and impractical to extract key frames from the enormous human demonstrations for more efficient data. Existing keyframe extraction techniques used in \cite{shridhar2022perceiver} failed to produce desirable results. 

\paragraph{Demonstration Generation.} We designed a motion planner for each task to generate demonstrations. We partitioned each task into sub-task stages for the planner. For each stage, we adopted the RMPflow controller \cite{cheng2020rmp} to plan motions toward keypoints. Unlike other approaches to data curation in simulation environments, this keypoint-based motion planner approach affords high sampling efficiency and facilitates imitation learning. While motion planning appeared to be challenging on particular tasks, as demonstrated in \cite{mu2021maniskill, gu2023maniskill2}, we introduced some prior design and practical techniques (details in Appendix~B of \cite{gong2023arnold}) to produce satisfactory outcomes. For example, we leveraged spherical linear interpolation (Slerp) to accommodate continuous manipulation in the \textsc{Cabinet} and \textsc{Water} tasks. As a result, our motion planner can efficiently generate demonstrations.

\paragraph{Augmentation With Human Annotations.} Despite the strength of motion planners, the diversity of produced demonstrations is highly dependent on the keypoints. To mitigate this problem, we collected about 2k human annotations of task configurations (\eg, object positions), which amount to considerably more diverse and higher quality data. Moreover, we augmented the data with additional relative positions and robot shifts to broaden data variations. Eventually, we curated demonstrations by running inference with ground-truth keypoints and verifying the validity of initial configurations in each execution. In total, we collected 10k valid demonstrations for the \benchmark benchmark (as in~\cref{tab:stats}), with each demonstration containing 4--6 keyframes.

\begin{table}[t!]
    \centering
    \caption{Dataset statistics. (1) \textit{Train}: training data. (2) \textit{Val}: validation data for model selection. (3) \textit{Test}: test data for i.i.d. evaluation. (4) \textit{Object}/\textit{Scene}/\textit{State}: \textit{Novel} splits for generalization evaluation. (5) \textit{State}$^*$: the \textit{Any State} split for generalization on arbitrary state.}
    \vspace{0.5em}
    \small
    \resizebox{\linewidth}{!}{
    \begin{tabular}{lrrrrrrrr}
        \toprule
         & \textit{Train} & \textit{Val} & \textit{Test} & \textit{Object} & \textit{Scene} & \textit{State} & \textit{State}$^*$ & Total \\
        \midrule
        \textsc{PickupObject} & 623 & 134 & 134 & 275 & 221 & 294 & 134 & 1,815 \\
        \textsc{ReorientObject} & 355 & 76 & 77 & 114 & 82 & 210 & 77 & 991 \\
        \textsc{OpenDrawer} & 554 & 119 & 119 & 155 & 255 & 348 & 119 & 1,669 \\
        \textsc{CloseDrawer} & 671 & 147 & 148 & 251 & 81 & 530 & 148 & 1,976 \\
        \textsc{OpenCabinet} & 319 & 69 & 69 & 81 & 181 & 241 & 69 & 1,029 \\
        \textsc{CloseCabinet} & 478 & 103 & 103 & 55 & 157 & 72 & 103 & 1,071 \\
        \textsc{PourWater} & 312 & 67 & 67 & 96 & 87 & 186 & 67 & 882 \\
        \textsc{TransferWater} & 259 & 56 & 56 & 51 & 50 & 119 & 56 & 647 \\
        \midrule
        Total & 3,571 & 771 & 773 & 1,078 & 1,114 & 2,000 & 773 & 10,080 \\
        \bottomrule
    \end{tabular}
    }
    \label{tab:stats}
\end{table}

\paragraph{Language Instructions.} For each demonstration, we sampled a template-based language instruction with our language generation engine. We designed several instruction templates with blanks for each task, and each template can be lexicalized with various phrase candidates. For example, the template ``\textit{pull the \textup{[position]} \textup{[object]} \textup{[percentage]} open}'' may be lexicalized into ``\textit{pull the top drawer $50\%$ open}''. In addition to the representation by explicit numbers, we also prepared a candidate pool of equivalent phrases (\eg, ``\textit{fifty percent}'', ``\textit{half}'', ``\textit{two quarters}'') for random replacement. Note that the instruction does not specify the initial state, so the agent must recognize the current state from the observation. We present template examples in Appendix~C of \cite{gong2023arnold}.

% \begin{table*}[t!]
% \centering
% \caption{Delexicalized instructions for all tasks. We sample each attribute during the generation process based on the proposed template. We increase diversity in our language descriptions for each trajectory by sampling from a variety of templates and applying paraphrasing techniques.}
% \vspace{0.5em}
% \begin{tabular}{lc}
% \toprule
% Task Name         &  Delexicalized Instructions \\
% \midrule
% PickObject        & raise [value\_object] [value\_height] above the ground. \\
% ReorientObject    & reorient [value\_object] [value\_degree] away from the up axis \\
% OpenCabinet       & open the [value\_position] [value\_object] [value\_percent] \\
% CloseCabinet      & close the [value\_position] [value\_object] [value\_percent] \\
% OpenDrawer        & open the [value\_position] [value\_object] [value\_percent] \\ 
% CloseDrawer       & close the [value\_position] [value\_object] [value\_percent]  \\ 
% PourWater         & pour [value\_percent] water out of [value\_object] \\
% TransferWater     & transfer [value\_percent] water to [value\_object] \\
% \bottomrule
% % TapWater          &<v><\text{o}><s> & fill the cup entirely full \\
% % \hline
% \end{tabular}
% \label{tab:template}
% \end{table*}

\subsection{Benchmark}\label{sec:benchmark}

\paragraph{Data Split.} Evaluating and improving the generalization abilities of robots is a major focus of \benchmark.
%\bx{the following paragraph needs to be clearer}
%\sy{Need clearer explanations of why we need demonstrations to do evaluation: multi-stage evaluation, providing start conditions for fair comparisons}
%\jy{why we take some demonstrations for evaluation: 1) verification; 2) first-stage oracle}
To this end, we randomly split the objects, scenes, and goal states into seen and unseen subsets, respectively. We then created the \textit{Normal} split by gathering data with seen objects, scenes, and states. The split was further shuffled and divided into \textit{Train}/\textit{Val}/\textit{Test} sets proportioned at $70\%/15\%/15\%$. Notably, in addition to providing valid initialization configurations, demonstrations for evaluation splits may be used to provide intermediate ground truth for diagnosing model performance (\cref{sec:ablation}).
% the validity of evaluation data instances and (2) diagnose models. 
Furthermore, we created the \textit{Generalization} splits \textit{Novel Object}/\textit{Scene}/\textit{State} by gathering data with one of the three components (\ie, objects, scenes, and goal states) unseen; \eg, the \textit{Novel Object} split comprises data of unseen objects, and seen scenes and states.

While the \textit{Novel State} split addresses the generalization of unseen goal states, we expect that grounding on continuous state representations should help the agent to adapt to any arbitrary state within a continuous range. Therefore, we make the \textit{Any State} split with seen objects and scenes, setting the goal states uniformly distributed over a continuous range, \eg, $0\%$--$100\%$. Such a design resembles universal tasks with arbitrary goal states and facilitates the evaluation of state generalization. \cref{tab:stats} presents the data statistics.

\paragraph{Metrics.} A task instance is regarded as a success when the success condition is satisfied continually for 2 seconds. The success condition requires the current state to be within a tolerance threshold from the goal state; \ie, the success range. The tolerances are derived according to human behaviors and are shown in \cref{tab:task_type}. Note that \textsc{TransferWater} imposes the extra condition that only $10\%$ or less of the water can be spilled. The execution of evaluation resembles the composition of sub-task stages in the motion planner (details in Appendix~B of \cite{gong2023arnold}). To avoid accidental triggering, we check the success condition after the agent completes the final stage. For example, in the task ``\textit{pour half of the water out of the cup}'', the agent succeeds if $40\% \sim 60\%$ of the water remains in the cup for 2 seconds after the agent has reoriented the cup upright. We have adopted success rate as the evaluation metric in the \benchmark.

\section{Experiments}\label{sec:exp}
\subsection{Experimental Setup}\label{sec:exp:setup}

\begin{table*}[t!]
\centering
\caption{Evaluation results of the models on various tasks and splits, measured by success rate and shown in percentages. The gray figures indicate performances with the first-phase ground truth. For each model, the first row shows the performance on the \textit{Test} set, and the following three rows show those on the \textit{Novel} splits of \textit{Object}, \textit{Scene}, and \textit{State}. The last row indicates the performances on the \textit{Any State} split. Tasks are abbreviated for more space. Average performances on eight tasks are appended to each row. \textbf{w/o L}: without language instruction. \textbf{$\dagger$}:~model variants with state modeling. \textbf{MT}: multi-task models.}
\vspace{4px}
\resizebox{\linewidth}{!}{
\begin{tabular}{lrrrrrrrrrrrrrrrrrr}
   \toprule
    \multicolumn{1}{c}{} & \multicolumn{2}{c}{\textsc{P.Object}} & \multicolumn{2}{c}{\textsc{R.Object}} & \multicolumn{2}{c}{\textsc{O.Drawer}} & \multicolumn{2}{c}{\textsc{C.Drawer}} & \multicolumn{2}{c}{\textsc{O.Cabinet}} & \multicolumn{2}{c}{\textsc{C.Cabinet}} & \multicolumn{2}{c}{\textsc{P.Water}} & \multicolumn{2}{c}{\textsc{T.Water}} & \multicolumn{2}{c}{Average} \\
    \midrule
    \textbf{6D-CLIPort} & 6.72 & \textcolor{NewGreen}{25.37} & 0.00 & \textcolor{NewGreen}{0.00} & 0.00 & \textcolor{NewGreen}{0.00} & 0.00 & \textcolor{NewGreen}{2.70} & 0.00 & \textcolor{NewGreen}{0.00} & 0.00 & \textcolor{NewGreen}{5.83} & 0.00 & \textcolor{NewGreen}{0.00} & 0.00 & \textcolor{NewGreen}{7.14} & 0.84 & \textcolor{NewGreen}{5.13}\\
    \object & 8.36 & \textcolor{NewGreen}{28.36} & 0.00 & \textcolor{NewGreen}{0.00} & 0.00 & \textcolor{NewGreen}{0.00} & 0.00 & \textcolor{NewGreen}{0.40} & 0.00 & \textcolor{NewGreen}{1.23} & 0.00 & \textcolor{NewGreen}{1.82} & 0.00 & \textcolor{NewGreen}{0.00} & 0.00 & \textcolor{NewGreen}{3.92} & 1.05 & \textcolor{NewGreen}{4.47} \\
    \scene & 10.41 & \textcolor{NewGreen}{24.43} & 0.00 & \textcolor{NewGreen}{0.00} & 0.00 & \textcolor{NewGreen}{1.57} & 0.00 & \textcolor{NewGreen}{0.00} & 0.00 & \textcolor{NewGreen}{0.55} & 1.27 & \textcolor{NewGreen}{1.27} & 0.00 & \textcolor{NewGreen}{0.00} & 0.00 & \textcolor{NewGreen}{12.00} & 1.46 & \textcolor{NewGreen}{4.98} \\
    \state & 0.00 & \textcolor{NewGreen}{0.00} & 0.00 & \textcolor{NewGreen}{0.00} & 0.00 & \textcolor{NewGreen}{0.57} & 0.75 & \textcolor{NewGreen}{1.13} & 0.00 & \textcolor{NewGreen}{0.83} & 0.00 & \textcolor{NewGreen}{2.78} & 0.00 & \textcolor{NewGreen}{0.00} & 0.00 & \textcolor{NewGreen}{16.81} & 0.09 & \textcolor{NewGreen}{2.77} \\
    \anystate & 10.45 & \textcolor{NewGreen}{29.10} & 1.30 & \textcolor{NewGreen}{2.60} & 0.84 & \textcolor{NewGreen}{0.00} & 0.68 & \textcolor{NewGreen}{1.35} & 0.00 & \textcolor{NewGreen}{5.80} & 0.00 & \textcolor{NewGreen}{2.91} & 0.00 & \textcolor{NewGreen}{1.49} & 0.00 & \textcolor{NewGreen}{7.14} & 1.66 & \textcolor{NewGreen}{6.30} \\
    \midrule
    \textbf{PerAct (w/o L)} & 25.37 & \textcolor{NewGreen}{33.58} & 14.29 & \textcolor{NewGreen}{7.79} & 17.65 & \textcolor{NewGreen}{36.13} & 47.30 & \textcolor{NewGreen}{52.03} & 8.70 & \textcolor{NewGreen}{34.78} & 7.77 & \textcolor{NewGreen}{10.68} & 14.93 & \textcolor{NewGreen}{11.94} & 5.36 & \textcolor{NewGreen}{14.29} & 17.67 & \textcolor{NewGreen}{25.15} \\
    \object & 29.09 & \textcolor{NewGreen}{26.55} & 8.77 & \textcolor{NewGreen}{3.51} & 3.87 & \textcolor{NewGreen}{20.00} & 24.70 & \textbf{\textcolor{NewGreen}{32.67}} & 0.00 & \textcolor{NewGreen}{0.00} & \textbf{1.82} & \textbf{\textcolor{NewGreen}{7.27}} & 16.67 & \textcolor{NewGreen}{29.17} & 9.80 & \textbf{\textcolor{NewGreen}{19.61}} & 11.84 & \textcolor{NewGreen}{17.35} \\
    \scene & 26.70 & \textcolor{NewGreen}{24.89} & 14.63 & \textcolor{NewGreen}{14.63} & 19.61 & \textcolor{NewGreen}{33.33} & 48.15 & \textcolor{NewGreen}{54.32} & \textbf{1.10} & \textcolor{NewGreen}{3.87} & 1.91 & \textcolor{NewGreen}{1.27} & 13.79 & \textcolor{NewGreen}{20.69} & 6.00 & \textcolor{NewGreen}{16.00} & 16.49 & \textcolor{NewGreen}{21.13} \\
    \state & 0.34 & \textcolor{NewGreen}{0.00} & 0.00 & \textcolor{NewGreen}{0.95} & 9.20 & \textcolor{NewGreen}{8.05} & 1.70 & \textcolor{NewGreen}{2.08} & 0.00 & \textcolor{NewGreen}{2.07} & 1.39 & \textcolor{NewGreen}{1.39} & 1.08 & \textcolor{NewGreen}{2.69} & 5.04 & \textbf{\textcolor{NewGreen}{8.40}} & 2.34 & \textcolor{NewGreen}{3.20} \\
    \anystate & 20.15 & \textcolor{NewGreen}{19.40} & 12.99 & \textcolor{NewGreen}{12.99} & 13.45 & \textcolor{NewGreen}{30.25} & 20.95 & \textcolor{NewGreen}{24.32} & \textbf{5.80} & \textbf{\textcolor{NewGreen}{26.09}} & \textbf{14.56} & \textbf{\textcolor{NewGreen}{15.53}} & 14.93 & \textcolor{NewGreen}{16.42} & 1.79 & \textcolor{NewGreen}{7.14} & 13.08 & \textcolor{NewGreen}{19.02} \\
    \midrule
    \textbf{PerAct} & 94.03 & \textbf{\textcolor{NewGreen}{97.76}} & 19.48 & \textcolor{NewGreen}{24.68} & 31.09 & \textcolor{NewGreen}{44.54} & \textbf{60.81} & \textcolor{NewGreen}{66.22} & \textbf{24.64} & \textcolor{NewGreen}{42.03} & 22.33 & \textcolor{NewGreen}{45.63} & \textbf{55.22} & \textcolor{NewGreen}{74.63} & \textbf{32.14} & \textcolor{NewGreen}{46.43} & 42.47 & \textcolor{NewGreen}{55.24} \\
    \object & 86.55 & \textbf{\textcolor{NewGreen}{92.73}} & \textbf{11.40} & \textbf{\textcolor{NewGreen}{35.09}} & \textbf{6.45} & \textcolor{NewGreen}{21.29} & \textbf{26.29}  & \textcolor{NewGreen}{27.89} & 0.00 & \textcolor{NewGreen}{0.00} & \textbf{1.82} & \textcolor{NewGreen}{5.45} & \textbf{36.46} & \textbf{\textcolor{NewGreen}{42.71}} & \textbf{13.73} & \textcolor{NewGreen}{13.73} & \textbf{22.84} & \textbf{\textcolor{NewGreen}{29.86}} \\
    \scene & \textbf{72.85} & \textbf{\textcolor{NewGreen}{84.62}} & \textbf{17.07} & \textcolor{NewGreen}{31.71} & 20.78 & \textcolor{NewGreen}{31.37} & \textbf{66.67} & \textcolor{NewGreen}{64.20} & 0.00 & \textbf{\textcolor{NewGreen}{4.97}} & 5.10 & \textcolor{NewGreen}{19.11} & \textbf{33.33} & \textcolor{NewGreen}{51.72} & 22.00 & \textbf{\textcolor{NewGreen}{36.00}} & \textbf{29.73} & \textcolor{NewGreen}{40.46} \\
    \state & 2.38 & \textcolor{NewGreen}{0.68} & 0.00 & \textcolor{NewGreen}{0.95} & \textbf{10.92} & \textcolor{NewGreen}{12.64} & \textbf{13.77} & \textcolor{NewGreen}{17.74} & 0.00 & \textcolor{NewGreen}{5.81} & 1.39 & \textcolor{NewGreen}{1.39} & 1.61 & \textcolor{NewGreen}{1.08} & \textbf{5.88} & \textcolor{NewGreen}{1.68} & 4.49 & \textcolor{NewGreen}{5.25} \\
    \anystate & 47.01 & \textbf{\textcolor{NewGreen}{50.75}} & 7.79 & \textbf{\textcolor{NewGreen}{19.48}} & \textbf{21.85} & \textcolor{NewGreen}{30.25} & 18.92 & \textcolor{NewGreen}{25.00} & \textbf{5.80} & \textcolor{NewGreen}{21.74} & 3.88  & \textbf{\textcolor{NewGreen}{15.53}} & 14.93 & \textcolor{NewGreen}{25.37} & 10.71 & \textcolor{NewGreen}{14.29} & 16.36 & \textcolor{NewGreen}{25.30} \\
    \midrule
    \textbf{PerAct$^\dagger$} & \textbf{94.78} & \textcolor{NewGreen}{95.52} & \textbf{24.68} & \textbf{\textcolor{NewGreen}{28.57}} & \textbf{36.13} & \textbf{\textcolor{NewGreen}{52.94}} & 60.14 & \textbf{\textcolor{NewGreen}{68.24}} & 23.19 & \textbf{\textcolor{NewGreen}{49.28}} & \textbf{30.10} & \textbf{\textcolor{NewGreen}{48.54}} & 49.25 & \textbf{\textcolor{NewGreen}{85.07}} & 28.57 & \textbf{\textcolor{NewGreen}{53.57}} & \textbf{43.36} & \textbf{\textcolor{NewGreen}{60.22}} \\
    \object & \textbf{87.27} & \textcolor{NewGreen}{91.27} & 10.53 & \textcolor{NewGreen}{32.46} & 1.94 & \textbf{\textcolor{NewGreen}{22.58}} & 18.73 & \textcolor{NewGreen}{25.10} & 0.00 & \textbf{\textcolor{NewGreen}{4.94}} & 0.00 & \textcolor{NewGreen}{5.45} & 34.38 & \textcolor{NewGreen}{33.33} & 9.80 & \textcolor{NewGreen}{11.76} & 20.33 & \textcolor{NewGreen}{28.36} \\
    \scene & 69.68 & \textcolor{NewGreen}{84.16} & 13.41 & \textcolor{NewGreen}{37.80} & \textbf{25.49} & \textbf{\textcolor{NewGreen}{38.43}} & 60.49 & \textbf{\textcolor{NewGreen}{67.90}} & 0.55 & \textbf{\textcolor{NewGreen}{4.97}} & 6.37 & \textbf{\textcolor{NewGreen}{19.75}} & 29.89 & \textbf{\textcolor{NewGreen}{63.22}} & \textbf{26.00} & \textcolor{NewGreen}{24.00} & 28.99 & \textbf{\textcolor{NewGreen}{42.53}} \\
    \state & 0.68 & \textcolor{NewGreen}{2.38} & 0.48 & \textcolor{NewGreen}{0.00} & 10.06 & \textcolor{NewGreen}{12.93} & 13.58 & \textbf{\textcolor{NewGreen}{18.11}} & 0.00 & \textbf{\textcolor{NewGreen}{6.22}} & 0.00 & \textcolor{NewGreen}{8.33} & 2.15 & \textcolor{NewGreen}{1.61} & \textbf{5.88} & \textcolor{NewGreen}{2.52} & 4.10 & \textcolor{NewGreen}{6.51} \\
    \anystate & \textbf{48.51} & \textcolor{NewGreen}{47.76} & \textbf{14.29} & \textcolor{NewGreen}{14.29} & 21.01 & \textbf{\textcolor{NewGreen}{33.61}} & \textbf{23.65} & \textbf{\textcolor{NewGreen}{28.38}} & 4.35 & \textcolor{NewGreen}{24.64} & 6.80 & \textcolor{NewGreen}{13.59} & \textbf{26.87} & \textbf{\textcolor{NewGreen}{31.34}} & \textbf{14.29} & \textbf{\textcolor{NewGreen}{19.64}} & \textbf{19.97} & \textbf{\textcolor{NewGreen}{26.66}} \\
    \midrule
    \textbf{PerAct (MT)} & 88.81 & \textcolor{NewGreen}{88.81} & 3.90 & \textcolor{NewGreen}{22.08} & 26.05 & \textcolor{NewGreen}{43.70} & 33.78 & \textcolor{NewGreen}{52.03} & 11.59 & \textcolor{NewGreen}{33.33} & 20.39 & \textcolor{NewGreen}{37.86} & 34.33 & \textcolor{NewGreen}{58.21} & 14.29 & \textcolor{NewGreen}{23.21} & 29.14 & \textcolor{NewGreen}{44.90} \\
    \object & 77.09 & \textcolor{NewGreen}{77.45} & 7.02 & \textcolor{NewGreen}{20.18} & 1.29 & \textcolor{NewGreen}{15.48} & 13.55 & \textcolor{NewGreen}{25.50} & 0.00 & \textcolor{NewGreen}{0.00} & \textbf{1.82} & \textbf{\textcolor{NewGreen}{7.27}} & 13.54 & \textcolor{NewGreen}{30.21} & 1.96 & \textcolor{NewGreen}{11.76} & 14.53 & \textcolor{NewGreen}{23.48} \\
    \scene & 68.78 & \textcolor{NewGreen}{77.83} & 10.98 & \textbf{\textcolor{NewGreen}{41.46}} & 13.33 & \textcolor{NewGreen}{25.88} & 29.63 & \textcolor{NewGreen}{55.56} & 0.00 & \textbf{\textcolor{NewGreen}{4.97}} & 5.73 & \textcolor{NewGreen}{9.55} & 19.54 & \textcolor{NewGreen}{43.68} & 4.00 & \textcolor{NewGreen}{16.00} & 19.00 & \textcolor{NewGreen}{34.37} \\
    \state & \textbf{9.18} & \textbf{\textcolor{NewGreen}{12.59}} & 0.00 & \textcolor{NewGreen}{1.90} & 8.05 & \textcolor{NewGreen}{12.07} & 9.43 & \textcolor{NewGreen}{13.40} & 0.00 & \textcolor{NewGreen}{2.07} & 1.39 & \textcolor{NewGreen}{0.00} & 2.69 & \textcolor{NewGreen}{6.45} & \textbf{5.88} & \textcolor{NewGreen}{5.88} & \textbf{4.58} & \textcolor{NewGreen}{6.80} \\
    \anystate & 36.57 & \textcolor{NewGreen}{45.52} & 5.19 & \textcolor{NewGreen}{10.39} & 15.13 & \textcolor{NewGreen}{19.33} & 16.22 & \textcolor{NewGreen}{21.62} & 1.45 & \textcolor{NewGreen}{8.70} & 0.97 & \textcolor{NewGreen}{3.88} & 8.96 & \textcolor{NewGreen}{14.93} & 3.57 & \textcolor{NewGreen}{17.86} & 11.01 & \textcolor{NewGreen}{17.78} \\
    \midrule
    \textbf{PerAct (MT)$^\dagger$} & 90.30 & \textcolor{NewGreen}{92.54} & 14.29 & \textcolor{NewGreen}{20.78} & 25.21 & \textcolor{NewGreen}{47.90} & 33.78 & \textcolor{NewGreen}{56.76} & 20.29 & \textcolor{NewGreen}{39.13} & 19.42 & \textcolor{NewGreen}{37.86} & 26.87 & \textcolor{NewGreen}{64.18} & 17.86 & \textcolor{NewGreen}{30.36} & 31.00 & \textcolor{NewGreen}{48.69} \\
    \object & 81.09 & \textcolor{NewGreen}{85.45} & 7.89 & \textcolor{NewGreen}{24.56} & 3.23 & \textbf{\textcolor{NewGreen}{22.58}} & 14.74 & \textcolor{NewGreen}{28.69} & 0.00 & \textbf{\textcolor{NewGreen}{4.94}} & \textbf{1.82} & \textcolor{NewGreen}{5.45} & 9.38 & \textcolor{NewGreen}{20.83} & 3.92 & \textbf{\textcolor{NewGreen}{19.61}} & 15.26 & \textcolor{NewGreen}{26.51} \\
    \scene & 67.87 & \textcolor{NewGreen}{79.64} & 7.32 & \textcolor{NewGreen}{29.27} & 12.94 & \textcolor{NewGreen}{25.49} & 39.51 & \textcolor{NewGreen}{65.43} & \textbf{1.10} & \textcolor{NewGreen}{4.42} & \textbf{7.01} & \textcolor{NewGreen}{14.01} & 10.34 & \textcolor{NewGreen}{43.68} & 8.00 & \textcolor{NewGreen}{22.00} & 19.26 & \textcolor{NewGreen}{35.49} \\
    \state & 2.04 & \textcolor{NewGreen}{3.06} & \textbf{0.95} & \textbf{\textcolor{NewGreen}{2.38}} & 9.20 & \textbf{\textcolor{NewGreen}{18.68}} & 6.98 & \textcolor{NewGreen}{11.13} & 0.00 & \textcolor{NewGreen}{3.73} & \textbf{2.78} & \textbf{\textcolor{NewGreen}{11.11}} & \textbf{6.45} & \textbf{\textcolor{NewGreen}{9.14}} & 1.68 & \textcolor{NewGreen}{4.20} & 3.76 & \textbf{\textcolor{NewGreen}{7.93}} \\
    
    \anystate & 46.27 & \textcolor{NewGreen}{47.01} & 12.99 & \textcolor{NewGreen}{12.99} & 12.61 & \textcolor{NewGreen}{23.53} & 14.86 & \textcolor{NewGreen}{26.35} & 4.35 & \textcolor{NewGreen}{5.80} & 4.85 & \textcolor{NewGreen}{8.74} & 16.42 & \textcolor{NewGreen}{25.37} & 3.57 & \textcolor{NewGreen}{5.36} & 14.49 & \textcolor{NewGreen}{19.39} \\
    \bottomrule
\end{tabular}
}
\label{tab:test_result}
\end{table*}

\paragraph{Models.} To evaluate the existing language-conditioned robotic manipulation models on \benchmark, we chose two \sota models as our primary focus: 6D-CLIPort~\cite{zheng2022vlmbench} and PerAct~\cite{shridhar2022perceiver}.
\begin{itemize}[leftmargin=*,noitemsep]
    \item 6D-CLIPort takes as input an RGB-D image from the top-down view and predicts end-effector poses for the current object and the target action. Each end-effector pose contains an action translation and a categorical prediction over discretized Euler angles. 6D-CLIPort comprises three branches to process the multi-modal input: Transporter-ResNet \cite{zeng2021transporter} for the spatial stream, CLIP visual encoder and language encoder \cite{radford2021learning} for the semantic stream. 
    % The learning objectives and optimization settings are the same with \cite{zheng2022vlmbench}, \eg, Adam optimizer at a learning rate of $1\times 10^{-3}$.
    \item PerAct takes RGB-D images as input to fuse a 3D voxelized grid. In addition, PerAct also requires the proprioception, including gripper states and the current timestep. The proprioception features are tiled on the voxel grid. Next, the hybrid voxel grid is downsampled and flattened to a sequence. Meanwhile, the language instruction is fed to a language encoder (\eg, CLIP \cite{radford2021learning}) and then appended to the sequence. PerAct uses Perceiver-IO~\cite{jaegle2021perceiver} to resample a compact latent representation from the multi-modal long sequence. After decoding, PerAct finally outputs a Q function over the original voxel grid for the prediction of action translation. Similar to 6D-CLIPort, PerAct also outputs a categorical distribution over discretized Euler angles for the prediction of action rotation. In contrast to the implementation in \cite{shridhar2022perceiver}, we discard the heads for predicting gripper and collision. Instead, we add an optional head for state prediction. 
    % Learning and optimization are the same with \cite{shridhar2022perceiver}.
\end{itemize}

Moreover, we considered three model variants of PerAct in our experiments: (1) PerAct without language (PerAct w/o L) for studying the importance of language-grounding, (2) PerAct with additional supervision on state value (PerAct$^\dagger$) to show the urgency of state modeling for tasks in \benchmark, and (3) PerAct trained in the multi-task setting (PerAct MT) given the great potential of multi-task learning shown in~\cite{shridhar2022perceiver}. For PerAct$^\dagger$, we provide additional state supervision by adding an extra output head to regress the normalized state values from the hidden features. We also tried other manipulation models; \eg, BC-Z in Appendix~D of \cite{gong2023arnold}.

% \begin{itemize}[nolistsep,noitemsep,leftmargin=*]
%     \item PerAct without language instruction (w/o L): We only apply this ablation on the single-task PerAct, since the language instruction is needed to distinguish tasks in multi-task setting.
%     \item A variant of PerAct with an extra output head for supervision on the state, which we refer to by the symbol $^\dagger$. During training, we normalize the state value into $[0,1]$ and regress this normalized state value from the final hidden features. This objective is inserted as part of the training loss. Despite the naive implementation, we hope the extra supervision could benefit the state understanding.
%     \item We evaluate the above two models as baselines under a single-task training scheme. Considering the feasibility of multi-task PerAct as demonstrated in \cite{shridhar2022perceiver}, we also conduct experiments on multi-task PerAct. On the other hand, to better probe the grounding of state, we add two model settings:
% \end{itemize}

\begin{figure}[t!]
    \centering
    \includegraphics[width=\linewidth]{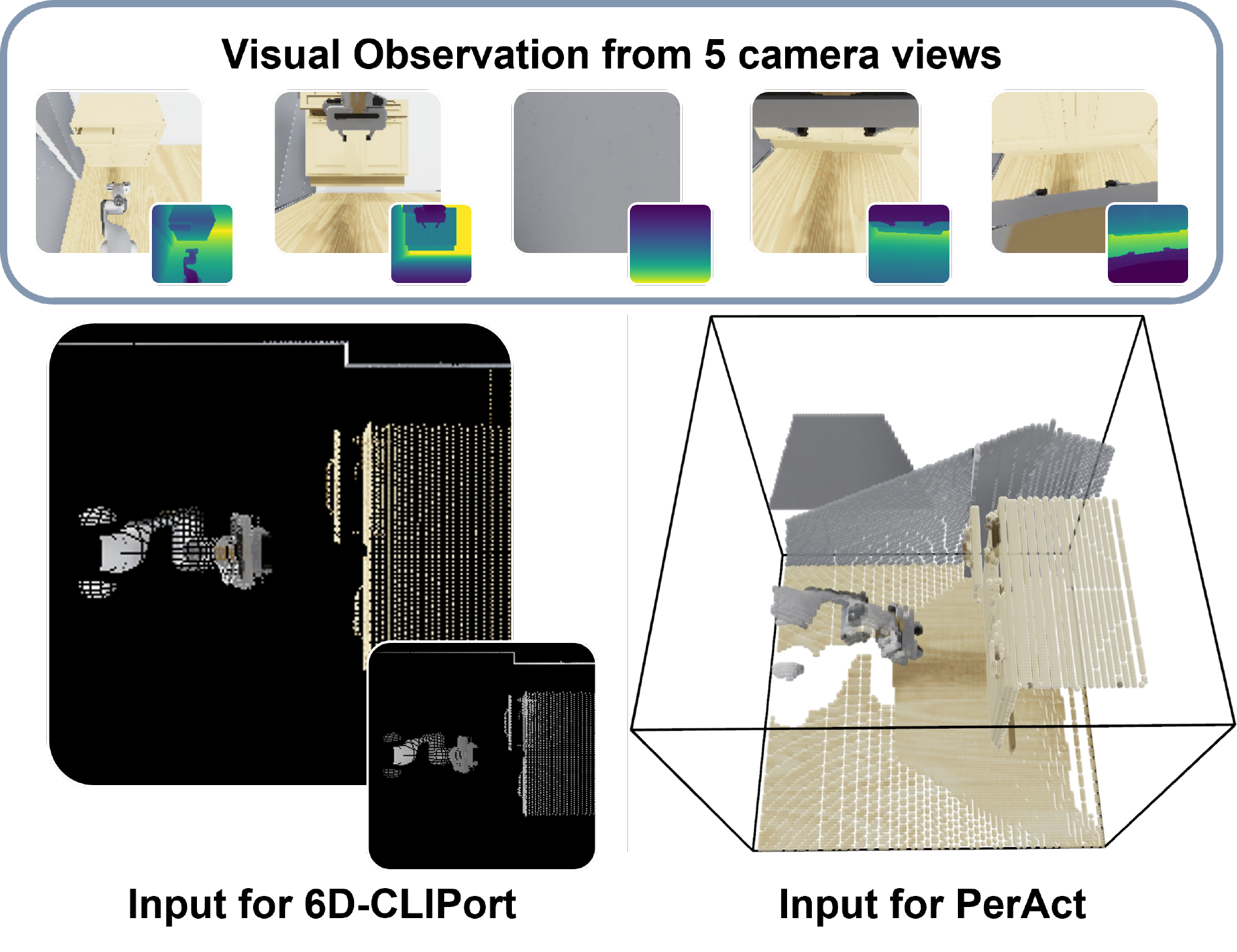}
    \caption{Visualization of the input representations of models. The five cameras are from the front, base, left, and two wrist views. The left camera view is occluded by the wall. Finally, these camera views fuse into an RGB-D image for 6D-CLIPort and a voxel grid for PerAct.}
    \label{fig:input}
\end{figure}

\paragraph{Implementation Details.} We obtain the visual representations for the models based on the five rendered RGB-D images as follows: With the camera parameters, we cast the pixels back to 3D and thus derive a point cloud for each view. With these point clouds in 3D scenes, we apply a perception bounding box to make the keypoint learning more tractable. In our setting, the cube spans $126$\,cm on each axis and the cube center $(p_x,p_y,p_z)$ is $50$\,cm away from the robot base along the robot's forward direction. 
% Hence, the scene context that models can perceive is included in the cube $(p_x\pm 63\text{cm}, p_y\pm 63\text{cm}, p_z\pm 63\text{cm})$. 
Next, we obtain the visual representations (visualized in \cref{fig:input}) as follows:
\begin{itemize}[leftmargin=*,noitemsep]
    \item For 6D-CLIPort, with each pixel occupying a size of 0.56\,cm, we can map the 126\,cm $\times$ 126\,cm perceived area to a $224\times 224$ top-down view image. We project the point clouds with their color and distance information onto this image. Note that the distance represents height rather than depth. If a pixel is related to multiple points (occlusion), the RGB values will be the average of these points and the distance will be the largest height.
    \item For PerAct, we set the size of voxel grid to be $120^3$, with each voxel covering the 3D context in a 1.05\,cm$^3$ volume. We then aggregate point clouds of all views into the voxel grid, including coordinates, RGB, positional embedding, and occupancy.
\end{itemize}
Please check \href{https://github.com/arnold-benchmark/arnold}{https://github.com/arnold-benchmark/arnold} for detailed implementations of the models and their variants.

\paragraph{Learning.} We trained all the models on the \textit{Train} split and performed model selection among ten checkpoints on the \textit{Val} split. Following common practice~\cite{zheng2022vlmbench,shridhar2022perceiver}, we used waypoints in demonstrations to facilitate model learning. Specifically, we unify the execution of tasks in \benchmark into a two-phase procedure: grasping the target object and then manipulating it toward the goal state. We followed the training settings of \cite{zheng2022vlmbench,shridhar2022perceiver}. Additionally, in multi-task training we sampled each training data by first uniformly sampling the task, then sampling a demonstration of the corresponding task. We set the number of training iterations to 100k for the single-task and 200k for the multi-task setting. We performed all the training on a single NVIDIA A100 GPU with batch size 8.

\paragraph{Evaluation Execution.} The evaluation executor of each task resembles the pipeline of a motion planner yet has a slight difference (see details in Appendix~B of \cite{gong2023arnold}). The model's predictions are converted back to keypoints of sub-task stages for evaluation execution. We make our evaluation strict and appropriate by avoiding shortcuts on the object state during robot motion. For example, the task instructing ``\textit{pull the cabinet $50\%$ open}'' will not be considered a success even if the cabinet is held around $50\%$ open for 2 seconds so long as the motion planner has not executed its final action. To eliminate the influence of a first-phase failure (\ie, failing to grasp the target object or object part), we conduct additional evaluations that provide first-phase ground truth.

\subsection{Experimental Results}
\label{sec:analysis}

% \zzh{in general I suggest we re-categorize this section. 
% This 4.2 Experimental Results section can contain only 3 parts: "Task-wise Results", "OOD Generalization", "Generalize to Arbitrary States". Discuss the two models at the beginning instead of making it a distinct subsection, since you will regardless discuss these two models when discussing the results.

% Then start another big section 4.3 called "Ablation Studies" and put all the rest into it, sub-categorizing into (1) "Input Ablation", which discusses "W/ VS W/o Language input", "W/ VS W/o Ground Truth states input";  (2) "Foundation Language Encoder Ablation".

%  Sim2Real section then becomes 4.4. 
 
%  It's easier for readers to digest when each section has not many subsections.}

We report experimental results in \cref{tab:test_result} (black) and present our findings and analyses below.

\paragraph{Across Models.} Comparing the baseline models 6D-CLIPort, PerAct, and PerAct (MT), we found that 6D-CLIPort fails on most of the tasks. We conjecture that this stems from (1) the information lost in the input representation when compressing complex 3D scenes into a single image and (2) the difficulty in regressing target height values for action translation. By contrast, the voxelized representations in PerAct provide rich 3D contexts that benefit model learning. 
With such differences, PerAct outperforms 6D-CLIPort significantly. Meanwhile, we observed performance drops for PerAct (MT) compared to PerAct on all tasks. This indicates that it is still difficult to leverage more diverse multi-task data for efficiently learning better task policies in \benchmark, especially given its challenges in both grounding and manipulation.

\paragraph{Across Tasks.} The most challenging tasks in \benchmark are \textsc{ReorientObject} and \{\textsc{Open},\textsc{Close}\}\textsc{Cabinet}. \textsc{ReorientObject} is difficult because it involves estimation of the bottle orientations, and models can often be confused by visually similar states. For example, the action for the goal state of $45\text{\textdegree}$ will lead to a goal state of $135\text{\textdegree}$ if the bottle orientation is reversed. Manipulating a cabinet proved to be challenging even with privileged information to specify goals~\cite{mu2021maniskill, gu2023maniskill2} as it requires accurate prediction of both interacting position, rotation, and precise continuous motion control. Replacing privileged information with instructions will only make it harder. Furthermore, we observed superior model performance in \textsc{PourWater} compared to \textsc{TransferWater}. This is because transferring water requires position alignment between cups to avoid spillage.

\paragraph{On Generalization Splits.} In general, we observed performance drops for most models when transferring to \textit{Novel} generalization splits, especially on the \textit{Novel State} split. This reveals that, without proper modeling of continuous states, generalizing the grounding of seen goal states to unseen ones remains challenging.
Meanwhile, the performance drop on the \textit{Novel Object} and \textit{Novel Scene} splits varies according to the tasks. For tasks where the objects occupy substantial space (\eg, drawer), the impact of unseen objects is more significant than unseen scenes. 

For the \textit{Any State} split, the models' performances were inferior compared with the \textit{Novel} \textit{Object}/\textit{Scene} split and superior to those on the \textit{Novel State} split. As goal states are uniformly sampled from a continuous spectrum, the success ranges of seen goal states are likely to cover a large portion of the spectrum, making the \textit{Any State} generalization interpolations of learned knowledge and skills. This suggests an interesting research question that can be investigated with \benchmark: How can the task learning model better generalize by interpolating within ranges and extrapolating to out-of-range goal states?

\paragraph{Remarks.} The key findings from our experiments with \benchmark are as follows:
\begin{itemize}[leftmargin=*,noitemsep,nolistsep]
\item Current models still struggle with tasks that require complex manipulation skills (\eg, manipulating cabinets). \textit{This heightens the demand for better policy learning models to tackle challenging manipulation tasks.}
\item The low success rate of models on all generalization splits motivates the necessity for (1) \textit{increasingly fine-grained representations for perceptual inputs}, (2) \textit{finer modeling of continuous object states}, and (3) \textit{better alignment between language and robot actions.}
\item The \textit{Any state} experiments suggest that \textit{state generalization could potentially be achieved through the interpolation of acquired knowledge and skills}. This promotes approaches with deeper insights into systematic generalization for robot skill adaptation.
\end{itemize}

\subsection{Ablation Studies}
\label{sec:ablation}

\paragraph{Influence of Language.} 
PerAct (w/o L), trained with a single-task scheme, exhibits a considerable performance gap behind PerAct. This indicates the importance of the goal-state information in \benchmark. Meanwhile, we observe a relatively small gap for \textsc{ReorientObject}. We believe this is due to (1) the bottleneck of this task lying in the ambiguity of visual perception, as discussed in \cref{sec:analysis} and (2) the difficulty of grounding angles from current visual observations. On the other hand, the significant performance gap on \textsc{PickupObject} shows the effectiveness of language grounding in visually more identifiable concepts such as translation distance.

\paragraph{Importance of State Modeling.} 
The PerAct variants with state supervision ($^\dagger$) were expected to realize a performance gain from explicit state prediction supervision. However, we observed marginal improvements on the \textit{Test} split and limited enhancements on generalization splits for this method. This indicates that such end-to-end state supervision is insufficient for state modeling in \benchmark and calls for better approaches to representing and modeling the continuous object states in robotic task learning.

\paragraph{With Intermediate Oracle.}
To better demonstrate how well models understand goal states, we provided models with first-phase ground truth (\ie, grasping positions) and report their performance in \cref{tab:test_result} (gray). As expected, we observed significant performance gains with such ground truth. Moreover, directly comparing the scores with the first-phase oracle, we can also observe larger gaps between PerAct and PerAct$^\dagger$, as well as PerAct (MT) and PerAct (MT)$^\dagger$. These results demonstrate that explicit state modeling is indeed beneficial for goal state understanding.

\begin{figure}[t!]
    \centering
    \includegraphics[width=\linewidth]{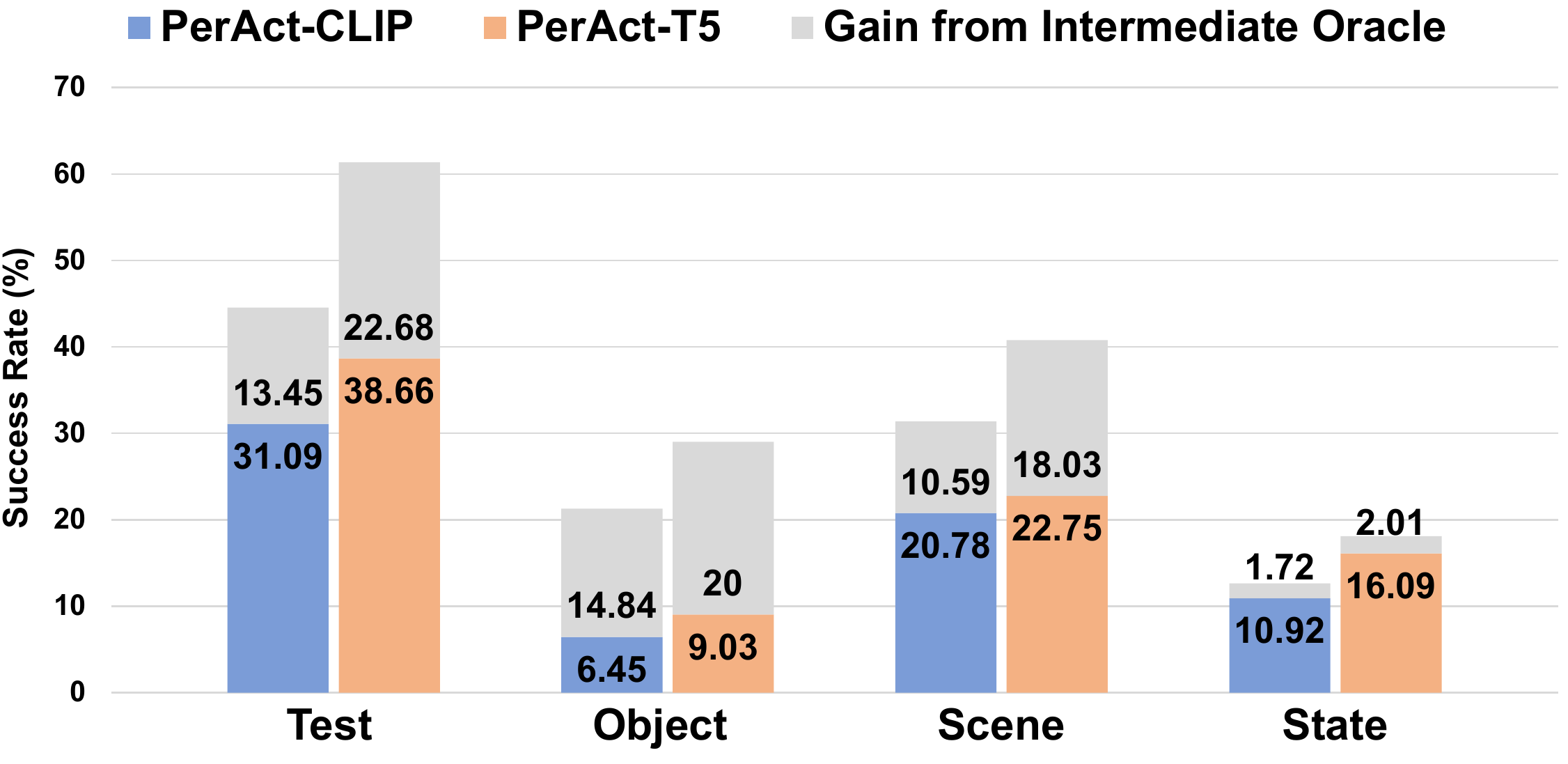}
    \caption{Model ablation results with different language encoders.}
    \label{fig:t5}
    % \vspace{-4mm}
\end{figure}

\paragraph{Choice of Language Encoder.} 
We investigated model ablation over the language encoder by switching the default language encoder CLIP~\cite{radford2021learning} in PerAct to T5-base~\cite{raffel2020exploring}. Due to space constraints, we report model performance only on \textsc{OpenDrawer}. As shown in \cref{fig:t5}, PerAct-T5 outperforms PerAct-CLIP on all the benchmarking splits. This may be due to the inefficacy of the global language representation learned in CLIP in representing fine-grained goal states for precise control. By contrast, T5 is a general-purpose language processing model that offers the ability to maintain detailed information through language modeling and generation. This shows that finer language embeddings may benefit the language grounding of robots to some extent.

% \paragraph{Discussion on state understanding.} \bx{We can discuss experiments in Tab.5 in a separate section as it's quite different from the generalization one, and arguing about not only skill generalization. The setting is different and will need a new setup paragraph for discussion.} Analysis of the results with partial ground truth. To further examine the capability on state generalization, we evaluate the models on the \textit{any state} split and report the results in \cref{tab:arbitrary_state}.

\subsection{Sim2Real Transfer Experiment}
\label{sec:exp:sim2real}

As a realistic simulation environment, one key question to address is: \textit{To what extent can the agents trained in \benchmark generalize to real-world scenarios}? To this end, we set up a real-world environment for testing the Sim2Real transfer capabilities of agents. Specifically, we used the Franka robot arm to manipulate previously unseen real-world objects with partial point cloud observations captured through a single RGB-D camera from the left view (\cref{fig:multi-view}). We experimented with PerAct, which was trained in \benchmark to open/close 2 different drawers and pick up 5 different objects. We mitigated the Sim2Real gap as follows:
\begin{enumerate}[leftmargin=*,noitemsep]
    \item Perception: We used high-fidelity rendering. Due to the imperfect depth sensor in the real world, we sprayed contrast aiding paint onto metallic and transparent areas. For diffuse objects with acceptable depth quality, further domain adaptation would be beneficial.
    \item Control: Instead of using the qpos control API, we used an inverse kinematics (IK) controller to perform real-robot actions, which avoids error accumulation.
\end{enumerate}

\begin{figure}[t!]
    \centering
    \includegraphics[width=\linewidth]{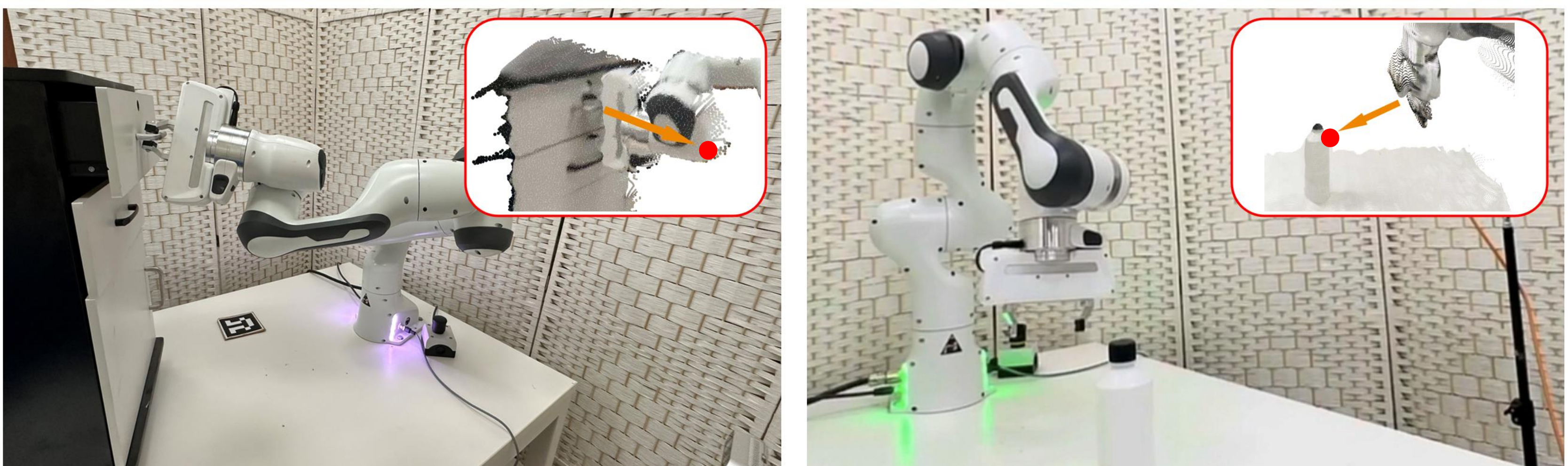}
    \vspace{-8pt}
    \caption{Real-world experiments with inference results shown on the upper right. The red dots indicate the predicted positions of the next action for the end-effector.}
    \label{fig:real-world}
    \vspace{-4mm}
\end{figure}

Additional experimental details can be found in Appendix~D of \cite{gong2023arnold}. Throughout our experiments, we observed that models trained in \benchmark show preliminary Sim2Real transfer capabilities; \ie, reasonable predictions for both picking up objects and manipulating drawers, as shown in~\cref{fig:real-world}. However, the actual robot manipulation continues to struggle because of the Sim2Real gap. For example, when opening a non-plastic drawer in the real world, the robot encounters high friction and is therefore susceptible to prediction errors that lead to inexecutable actions (\eg, exceeding the critical friction angle). With more fine-grained object assets, we believe that the flexible design of the \benchmark simulator can gradually close this Sim2Real gap by providing more realistic simulations. 

% compared with Peract

% Both the multi-task model and the single-task model can achieve some level of success in picking up objects and closing drawers. However, when it comes to opening a non-plastic drawer in the real world, the robot encounters high friction, making it difficult to move. If the robot's prediction is even slightly off, it may exceed the critical friction angle and overload the robot arm. 

% Therefore, despite the decent inference results shown in Figure \ref{fig:real-world}, the robot still fails to open the real-world drawer correctly. Additionally, the model may occasionally manipulate the object to the intended state, but if the instruction changes to a different state description, the model fails to achieve the desired state. These findings suggest that more research is necessary for Sim2Real transfer for language-conditioned manipulation tasks. We provide more details in \supp.
%respectively under 19, 8 and 31 different settings (object poses, camera poses, instructions, \etc), and achieved success in 6, 0 and 4 cases respectively. 

% The following remarks summarize the difficulties that we encountered and suggest avenues for future work:
% \begin{itemize}[leftmargin=*,noitemsep,nolistsep]

% \item The control gap is rather huge; for example, our model  performs well in the opening drawer task in the simulator. In the real world, however, although the inference can still yield good results, opening the real-world drawer must overcome potentially high friction. If the prediction deviates slightly, the critical friction angle may be reached, resulting in an overloading of the robot arm. %Other work \cite{shridhar2022perceiver, yang2022let} use a plastic drawer to bypass this issue.

% \item We use a single camera to acquire the point cloud, but we found that training the model using multiple cameras also enables it to work satisfactorily and complete tasks such as picking up objects.

% \item We found that the inference of continuous goals in the real world remains unsatisfactory; for example, an instruction can give the correct action output, but changing the task goal for the same task may produce a large deviation and leads to task failure (not matching the state description).
% \end{itemize}
% \vspace{-4mm}

\section{Conclusion}

We have presented \benchmark, a benchmark for language-grounded task learning in realistic 3D interactive environments with diverse scenes, objects, and continuous object states. We devised a systematic benchmark comprising eight challenging language-grounded robot tasks and evaluation splits for robot skill generalization in novel scene, object, and goal-state scenarios. We conducted extensive experiments and analyses to pinpoint the limitations of the current models and identified promising research directions for grounded task learning.

\paragraph{Limitations and Future Work.} (1) Despite our focus on realistic simulation, the gap between \benchmark and real-world scenarios still remains. However, with our flexible design, we can reduce this gap by adding more realistic variations of both scenes and objects \cite{geng2022gapartnet} to the assets library. (2) Current tasks in \benchmark do not require much high-level planning knowledge. However, as we found in experiments, short-horizon fine-grained control has not been well solved yet. This defers long-horizon tasks to the future. (3) Despite the diversity we inject in language, the template-based generation inevitably restricts language variations. As a key component for generalization, it is critical to extend \benchmark with richer language instructions (\eg, by prompting LLMs) in future work. (4) Current imitation learning models rely on prior simplifications (\eg, the two-phase learning) to learn sparse keypoints. Limited by the amount and diversity of expert policies, we need improved methods for modeling continuous object states and learning generalizable policies from scarce data. (5) The realistic and resourceful environment provided by \benchmark can also facilitate the pursuit of versatile agents with comprehensive capabilities \cite{jiang2022vima,huang2022perceive}. (6) Scaling up demonstrations is another crucial future step, which can induce stronger capabilities \cite{brohan2022can,brohan2022rt}.

\section*{Acknowledgments}
This work is supported in part by the National Key R\&D Program of China (2021ZD0150200) and NVIDIA GPU Grant. In addition, we thank Chiao Lu for the help with the environment setup for large-scale experiments. 

%\nocite{duan2022survey,deitke2022retrospectives}

{\small
\bibliography{ref}
\bibliographystyle{IEEEtranS}
}
\clearpage
\appendix
\section{Environment}
\subsection{Assets in USD Format}

Universal Scene Description (USD) is a format for 3D scene descriptions. The process of parsing assets into Omniverse involves the transformation of graphical data into USD files. This primarily provides developers with a streamlined method of accessing and retrieving relevant information from the assets, including scene files, articulation bodies, animations, \etc. By utilizing a USD file format, users can easily access a wide range of assets and subsequently inform Omniverse of the relevant content required for their application, eliminating the need for tedious, manual retrieval of information from each asset individually.
 
\paragraph{Working with the 3D-Front dataset.} The 3D-Front dataset consists of a set of professionally designed synthetic indoor scenes. It features a large number of rooms that are furnished with high-quality textured 3D models. The dataset is composed of three primary components: models, scenes, and textures. And it contains tens of thousands of room layouts with thousands of furnished objects. To parse the 3D-Front dataset into USD format, we apply the following steps:
\begin{itemize}[nolistsep,leftmargin=*]
  \item Parse the original scene files (JSON) into a data frame containing the mesh and furniture information.
  \item Use Maya MEL script to load scenes into Autodesk Maya.
  \item Apply Maya and Omniverse converter to save the scenes in USD format.
\end{itemize}

\begin{figure*}[t!]
  \centering
  \includegraphics[width = \linewidth]{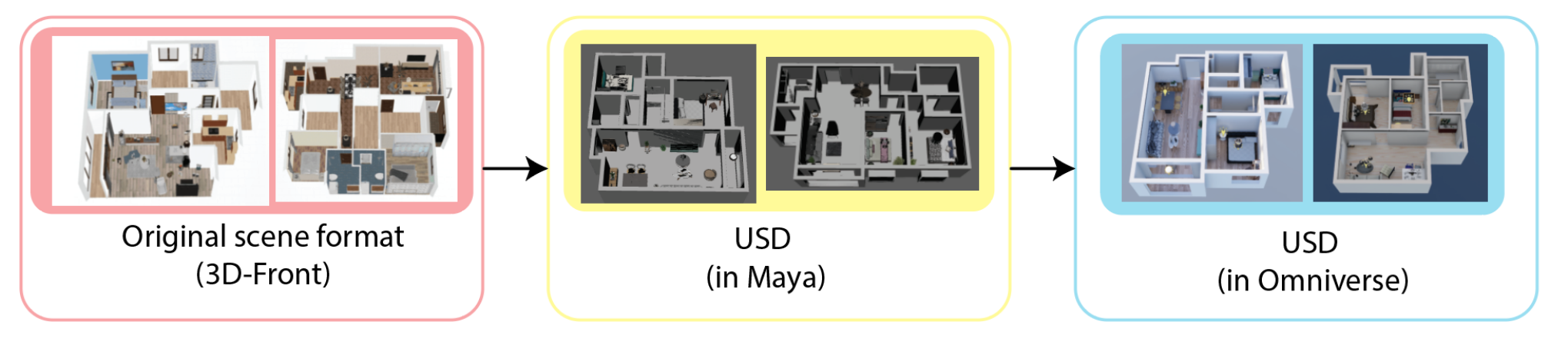}
  \caption{Pipeline of scene parsing. After pre-processing the original 3D-Front scene data, we build an automatic pipeline to load the scene layout into Autodesk Maya with custom designs. Next, we convert the layout file to USD format and deploy it in Omniverse.}
  \label{fig:app_scene}
\end{figure*}
 
 \paragraph{Working with articulated bodies.} The articulated bodies in \benchmark mainly come from SAPIEN. We use built-in tools of Omniverse Isaac Sim to parse the original articulated bodies (URDF) into USD format.
 
\begin{figure}[t!]
  \centering
  \includegraphics[width = \linewidth]{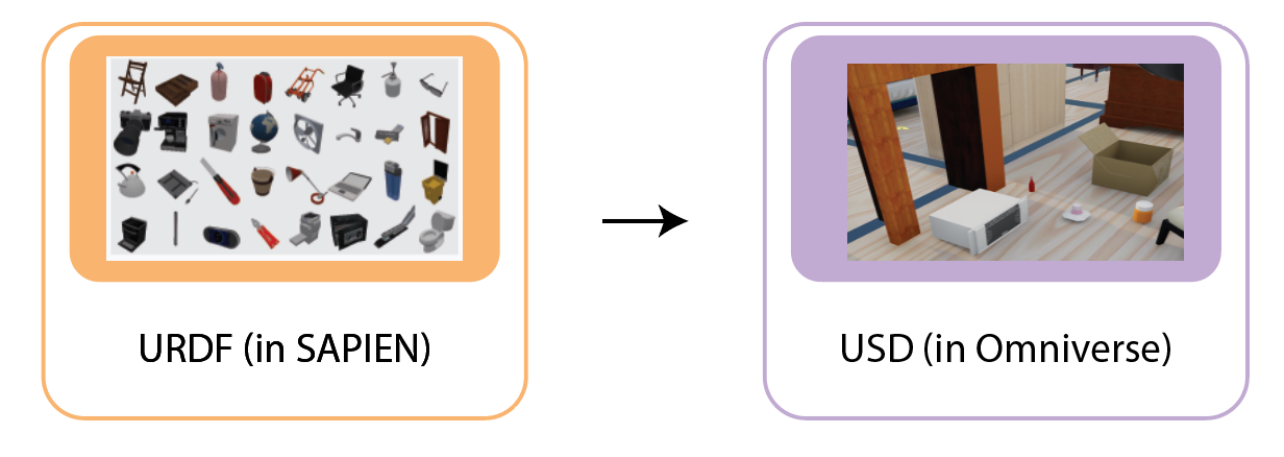}
  \caption{Pipeline of parsing articulated bodies. To convert the original articulated bodies into USD format, we refine the built-in functionalities in Omniverse Isaac Sim and modify the assets manually.}
  \label{fig:app_urdf}
\end{figure}

\subsection{Speed}
With five $128 \times 128 $ cameras, our system runs at 17 fps for liquid simulation and 37 fps for rigid body simulations on an NVIDIA RTX 3090 GPU and AMD 5950X CPU. Currently, Omniverse Isaac Sim only supports serial cameras. We expect a huge performance improvement with the upcoming release of Omniverse Isacc Sim which supports parallel cameras.

\subsection{Randomness}
\begin{figure}
\centering
    \begin{subfigure}[b]{\linewidth}
         \centering
         \includegraphics[width=\linewidth]{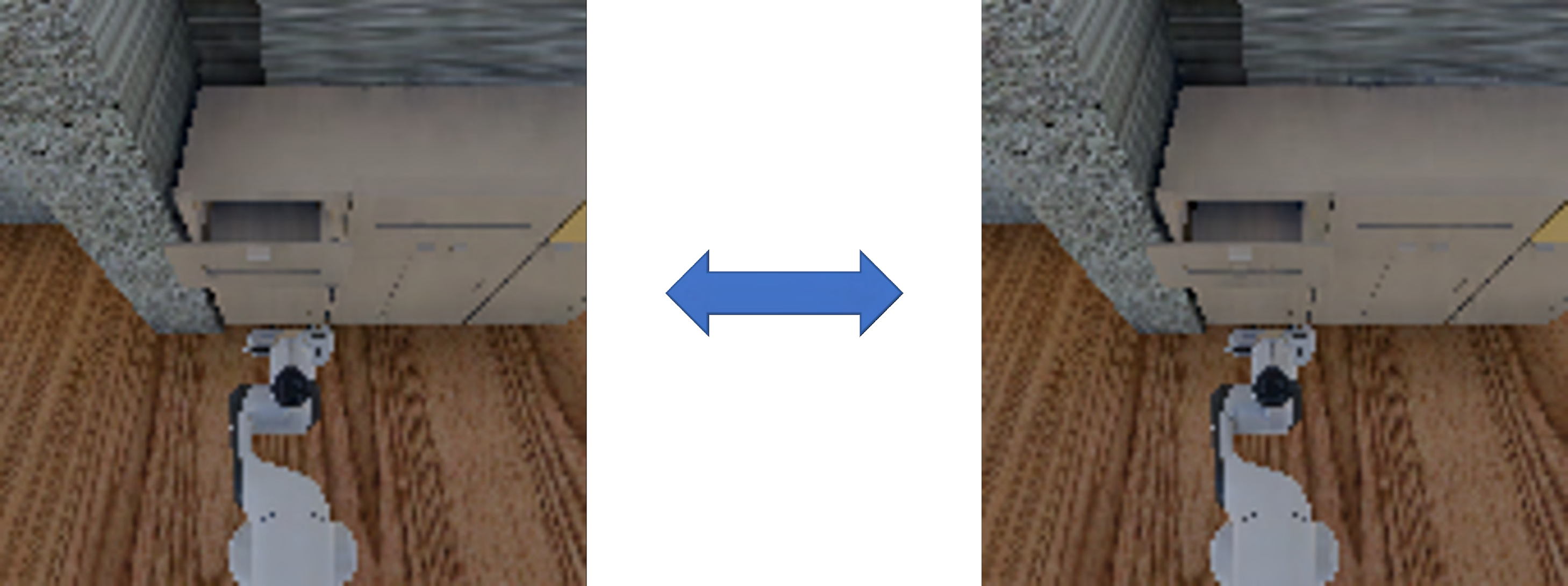}
         \caption{Two rendered images with a subtle difference.}
     \end{subfigure}
     \hfill
     \vspace{3pt}
     \begin{subfigure}[b]{\linewidth}
         \centering
         \includegraphics[width=0.8\linewidth]{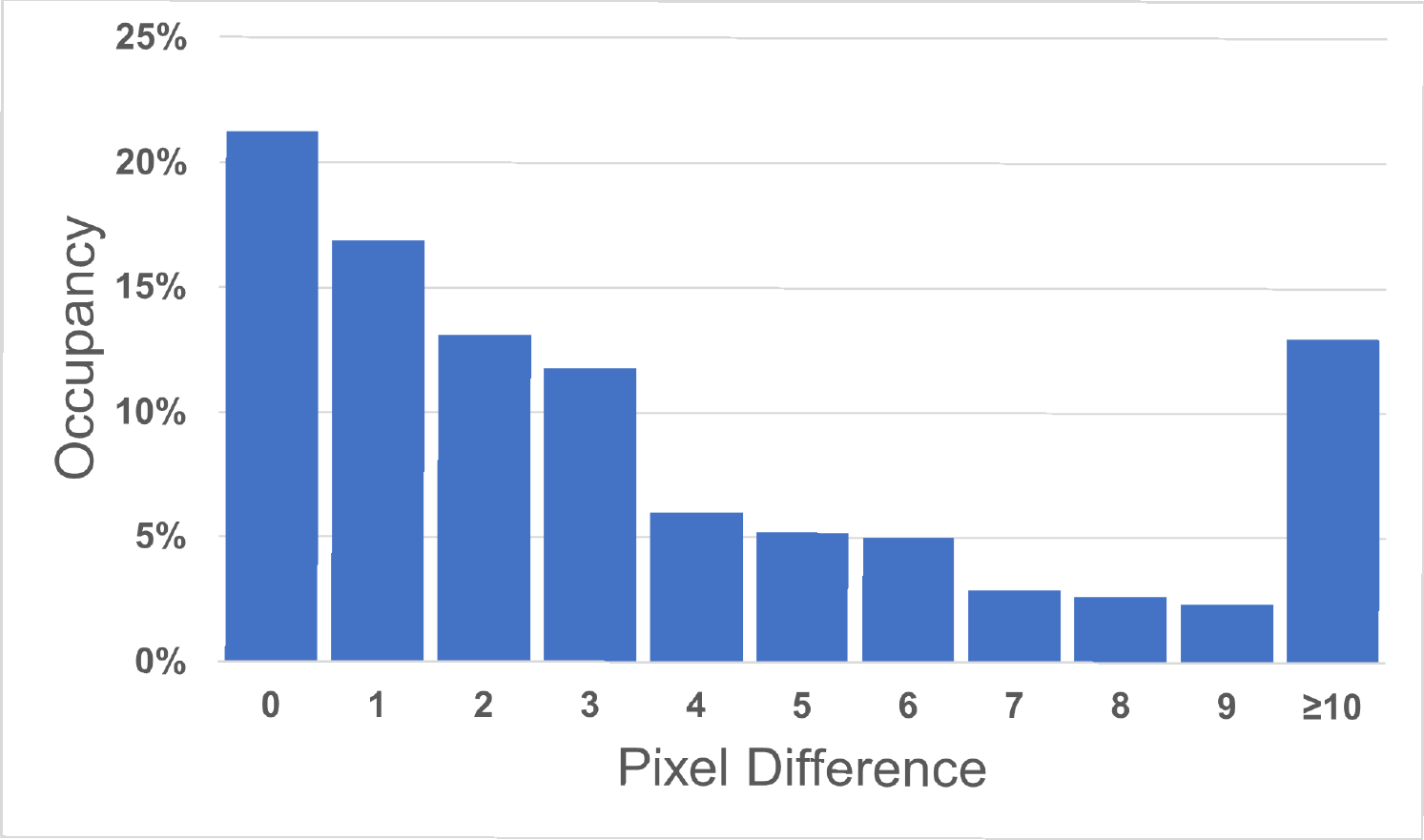}
             \caption{Distribution of pixel difference.}
     \end{subfigure}
    \caption{An example that shows the rendering randomness in \benchmark. (a) Two $128\times 128$ images rendered from the same frame exhibit a subtle difference, with about $78.8\%$ of the pixels being different. (b) The distribution of pixel difference. Here the difference per pixel is measured by the sum of RGB differences. More than $50\%$ of the pixels differ less than a value of $3$.}
    \label{fig:rendering_noise}
\end{figure}

Different from previous work \cite{zheng2022vlmbench, shridhar2022perceiver}, the rendering randomness in \benchmark would result in non-deterministic images. This means two images rendered from the same frame may exhibit a subtle difference. \cref{fig:rendering_noise} shows an example where about $78.8\%$ of the pixels are different, together with the distribution of pixel difference.

\section{Task Details}\label{sec:task_details}
\begin{figure*}
\centering
    \includegraphics[width=0.8\linewidth]{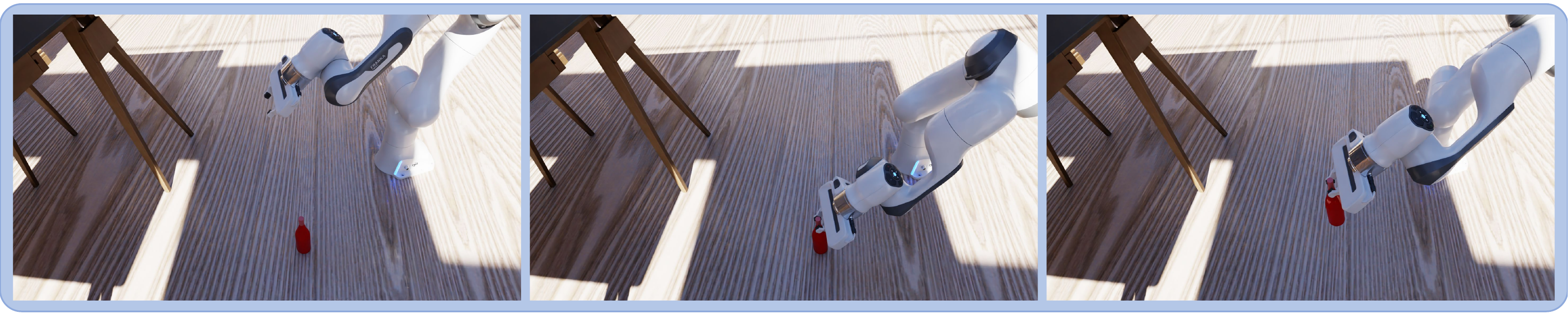}
    \includegraphics[width=0.8\linewidth]{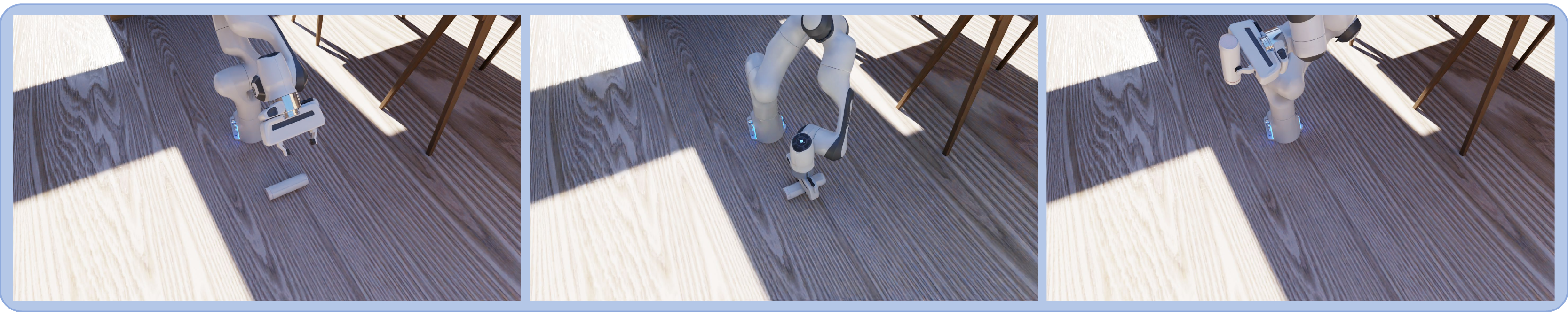}
    \includegraphics[width=0.8\linewidth]{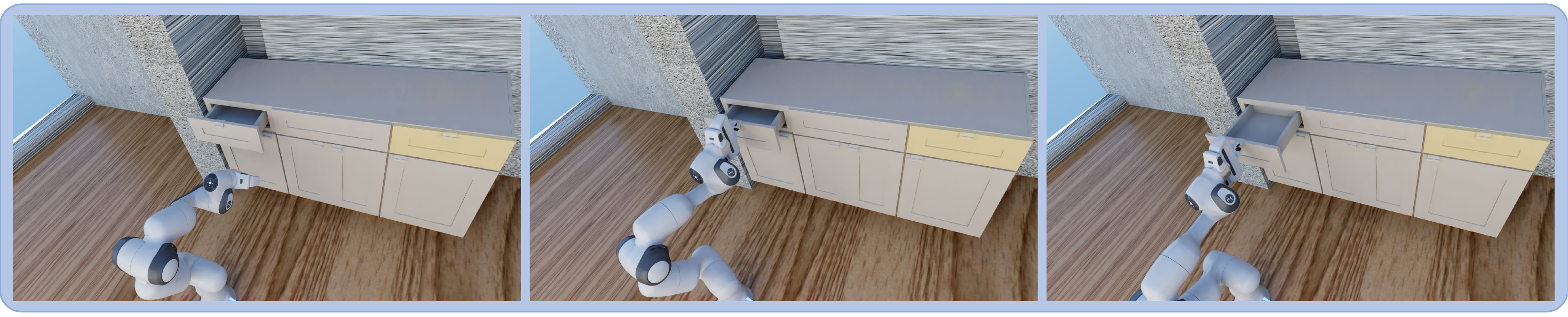}
    \includegraphics[width=0.8\linewidth]{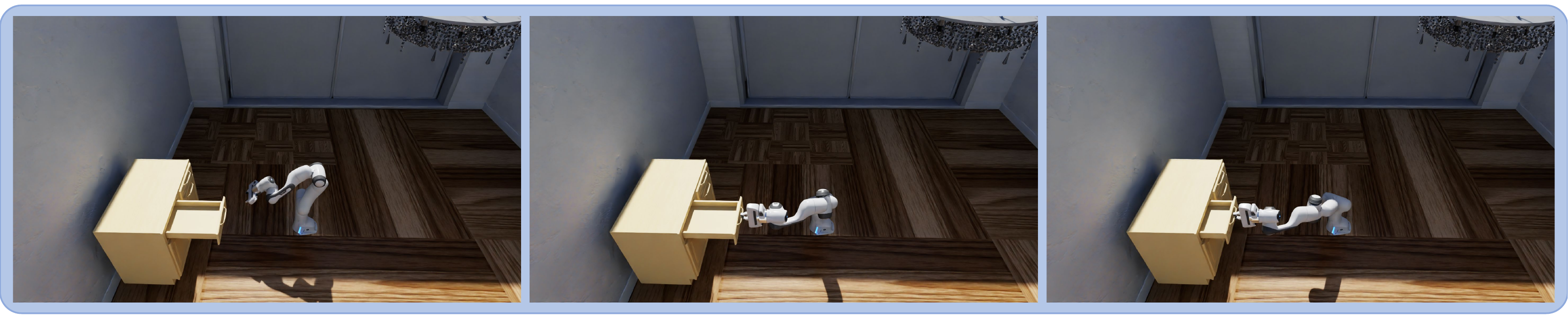}
    \includegraphics[width=0.8\linewidth]{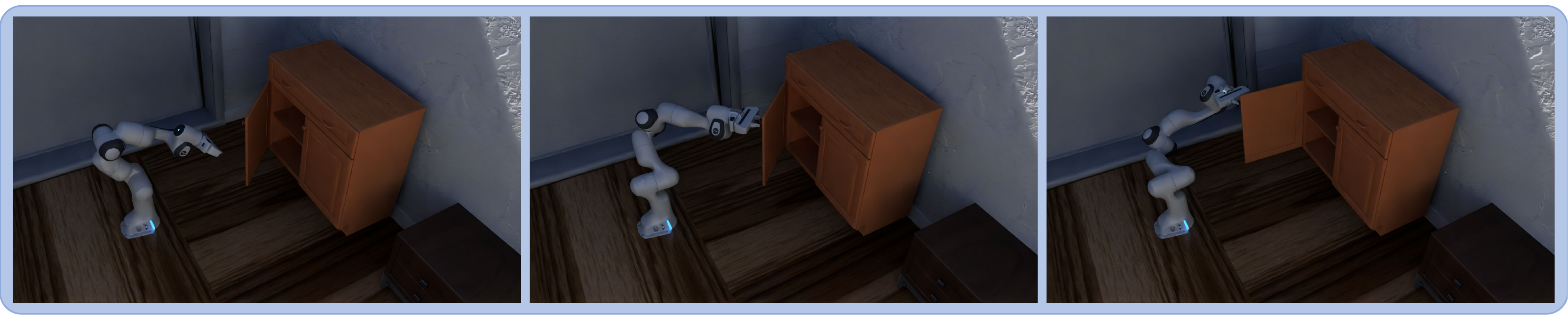}
    \includegraphics[width=0.8\linewidth]{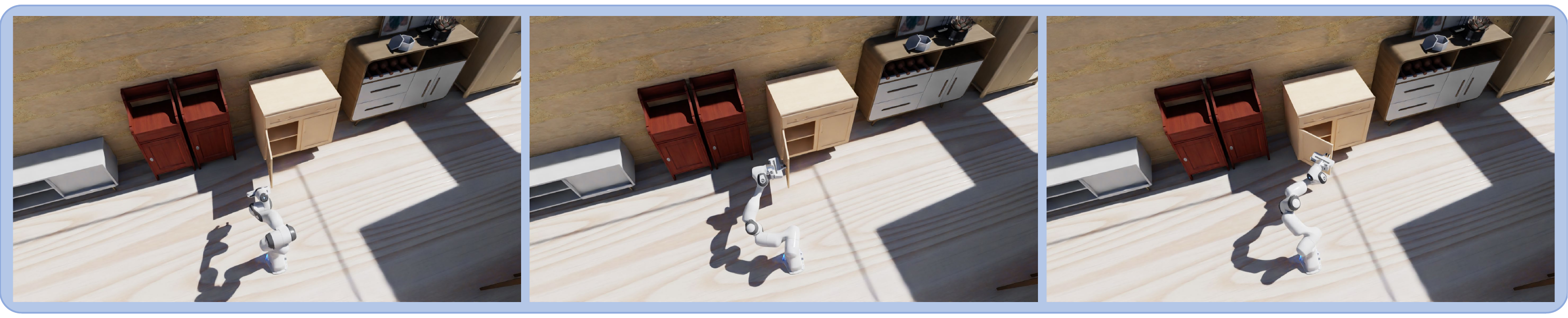}
    \includegraphics[width=0.8\linewidth]{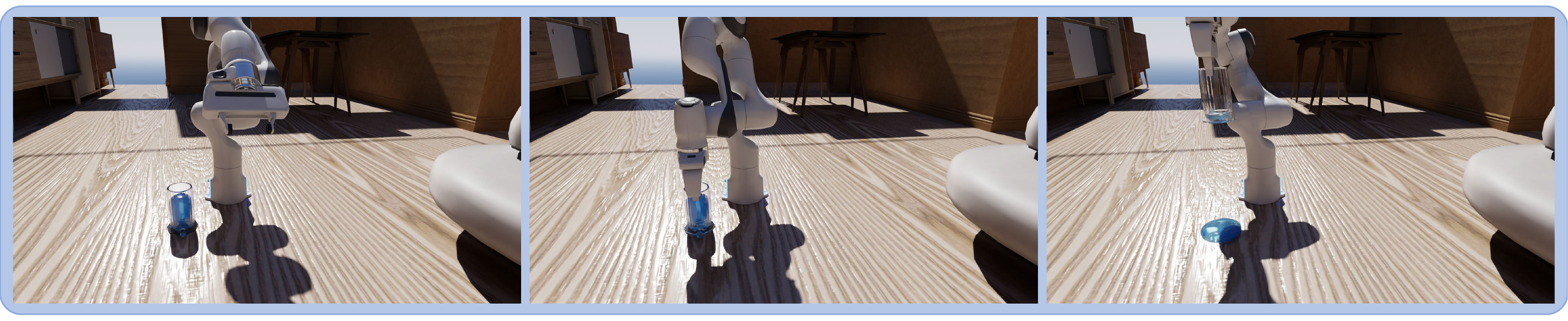}
    \includegraphics[width=0.8\linewidth]{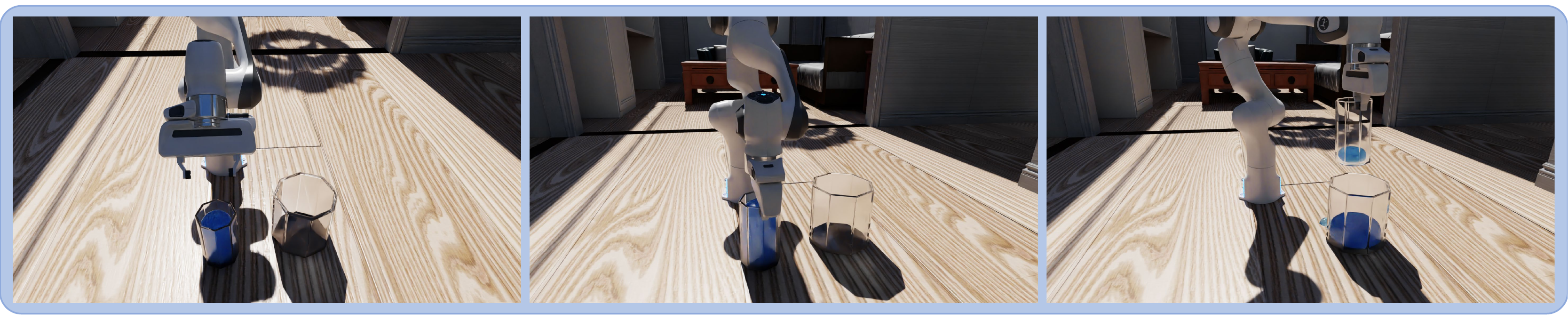}
    \caption{Illustrations of the 8 tasks in \benchmark.}
    \label{fig:illustration}
\end{figure*}

\begin{figure}[t!]
    \centering
    \includegraphics[width = \linewidth]{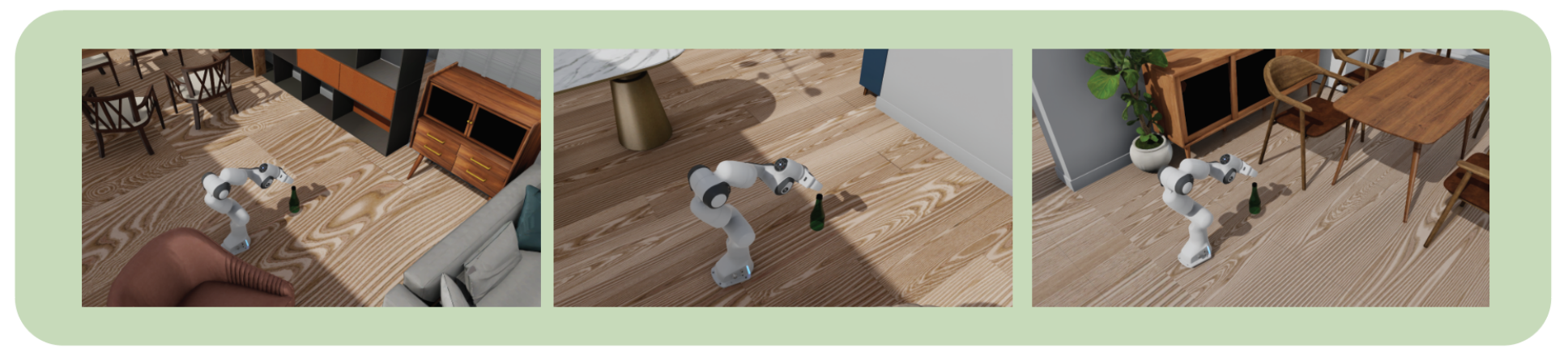}
    \caption{Scene variations.}
    \label{fig:scenes}
\end{figure}

\begin{figure}[t!]
    \centering
    \includegraphics[width = \linewidth]{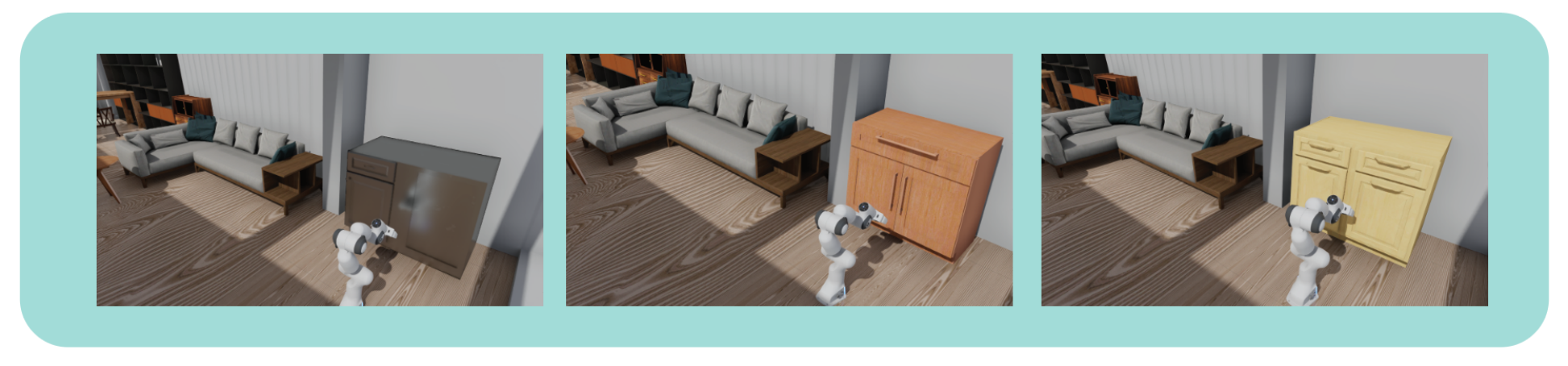}
    \caption{Object variations.}
    \label{fig:obj_var}
\end{figure}

\begin{figure}[t!]
    \centering
    \includegraphics[width = \linewidth]{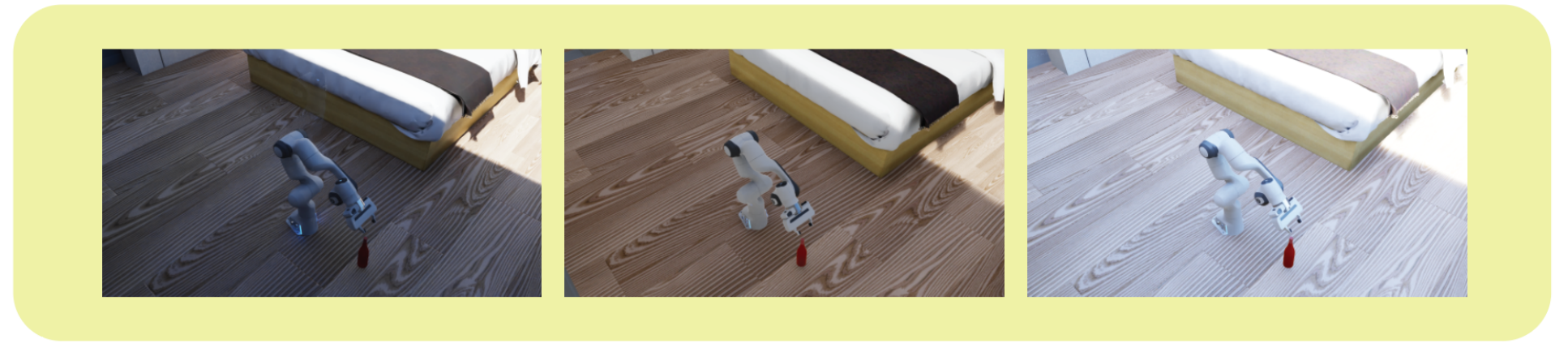}
    \caption{Lighting variations.}
    \label{fig:light}
\end{figure}

\begin{figure}[t!]
    \centering
    \includegraphics[width = \linewidth]{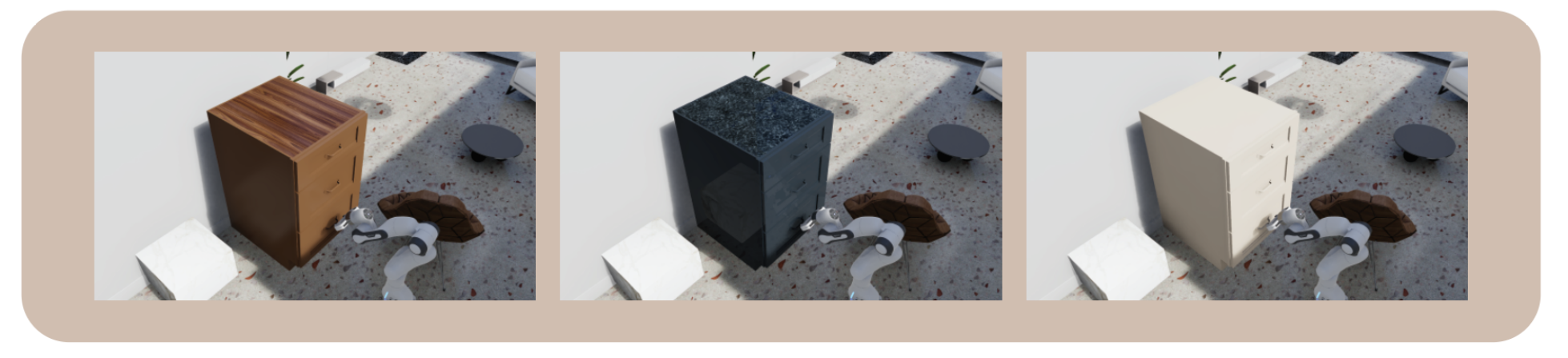}
    \caption{Material variations.}
    \label{fig:obj_mat}
\end{figure}

We provide illustrative examples of some tasks in \cref{fig:illustration}. \cref{fig:scenes,fig:obj_var,fig:light,fig:obj_mat} illustrate the variations from different aspects in \benchmark. In this section, we provide the implementation details of each task. In our notations, we use $o$ to denote visual observation, $a$ for action (\ie, position and rotation with regard to the world). $\delta$ is a function to compute the pre-grasp pose.

% \begin{figure}[t!]
%     \centering
%     \includegraphics[width = \linewidth]{figures/continous_goal.PNG}
%     \caption{Continous control and goal state.}
%     \label{fig:continuous}
% \end{figure}

\subsection{\textbf{\textsc{PickupObject}}}
\paragraph{Motion planner.} The motion planner of \textsc{PickupObject} consists of four sub-task stages. We depict each stage as follows, beginning with the visual observation and ending with a consequent end effector pose.
\begin{enumerate}[noitemsep]
    \item Observe $o_1$. Move the end effector to the pre-grasp pose. Reach $a_1=\delta(a_2)$.
    \item Observe $o_2$. Move the end effector to reach the position for grasping. Reach $a_2$.
    \item Observe $o_3$. Close gripper. Reach $a_3$.
    \item Observe $o_4$. Lift the object to the goal height. Reach $a_4$.
\end{enumerate}

\paragraph{Learning and evaluation.} We extract two observation-action pairs for two-phase learning: $o_1,a_2$ and $o_4,a_4$. During evaluation, the robot executes the action predictions $\tilde{a}_2$ and $\tilde{a}_4$ similar to the motion planner:
\begin{enumerate}[noitemsep]
    \item Move to $\delta(\tilde{a}_2)$ for pre-grasp.
    \item Move to $\tilde{a}_2$ and close gripper.
    \item Move to $\tilde{a}_4$ for goal state.
\end{enumerate}

\subsection{\textbf{\textsc{ReorientObject}}}
\paragraph{Motion planner.} The motion planner of \textsc{ReorientObject} consists of four sub-task stages. We depict each stage as follows, beginning with the visual observation and ending with a consequent end effector pose.
\begin{enumerate}[noitemsep]
    \item Observe $o_1$. Move the end effector to the pre-grasp pose. Reach $a_1=\delta(a_2)$.
    \item Observe $o_2$. Move the end effector to reach the position for grasping. Reach $a_2$.
    \item Observe $o_3$. Close gripper. Reach $a_3$.
    \item Observe $o_4$. Reorient the object to goal orientation. Reach $a_4$.
\end{enumerate}

\paragraph{Learning and evaluation.} We extract two observation-action pairs for two-phase learning: $o_1,a_2$ and $o_4,a_4$. During evaluation, the robot executes the action predictions $\tilde{a}_2$ and $\tilde{a}_4$ similar to the motion planner:
\begin{enumerate}[noitemsep]
    \item Move to $\delta(\tilde{a}_2)$ for pre-grasp.
    \item Move to $\tilde{a}_2$ and close gripper.
    \item Rotate to $\tilde{a}_4$ for goal state.
\end{enumerate}

\subsection{\textbf{\textsc{OpenDrawer} and \textsc{CloseDrawer}}}
\paragraph{Motion planner.} The motion planner for these two tasks consists of four sub-task stages. We depict each stage as follows, beginning with the visual observation and ending with a consequent end effector pose.
\begin{enumerate}[noitemsep]
    \item Observe $o_1$. Move the end effector to the pre-grasp pose. Reach $a_1=\delta(a_2)$.
    \item Observe $o_2$. Move the end effector to reach the position for grasping. Reach $a_2$.
    \item Observe $o_3$. Close gripper. Reach $a_3$.
    \item Observe $o_4$. Gradually pull or push the drawer until the goal condition is satisfied. We apply linear interpolation on the action translation to slow down since excessive movement will result in the detachment of gripper. Reach $a_4$.
\end{enumerate}

\paragraph{Learning and evaluation.} We extract two observation-action pairs for two-phase learning: $o_1,a_2$ and $o_4,a_4$. During evaluation, the robot executes the action predictions $\tilde{a}_2$ and $\tilde{a}_4$ similar to the motion planner:
\begin{enumerate}[noitemsep]
    \item Move to $\delta(\tilde{a}_2)$ for pre-grasp.
    \item Move to $\tilde{a}_2$ and close gripper.
    \item Gradually interpolate like the motion planner toward $\tilde{a}_4$ for goal state.
\end{enumerate}

\subsection{\textbf{\textsc{OpenCabinet} and \textsc{CloseCabinet}}}
\paragraph{Motion planner.} The motion planner for these two tasks consists of four sub-task stages. We depict each stage as follows, beginning with the visual observation and ending with a consequent end effector pose.
\begin{enumerate}[noitemsep]
    \item Observe $o_1$. Move the end effector to the pre-grasp pose. Reach $a_1=\delta(a_2)$.
    \item Observe $o_2$. Move the end effector to reach the position for grasping. Reach $a_2$.
    \item Observe $o_3$. Close gripper. Reach $a_3$.
    \item Observe $o_4$. Gradually pull or push the cabinet until the goal condition is satisfied. We apply linear interpolation on the action translation to slow down, and apply spherical linear interpolation (Slerp) on the action rotation to meet the revolute constraint. Reach $a_4$.
\end{enumerate}

\paragraph{Learning and evaluation.} We extract two observation-action pairs for two-phase learning: $o_1,a_2$ and $o_4,a_4$. During evaluation, the robot executes the action predictions $\tilde{a}_2$ and $\tilde{a}_4$ similar to the motion planner:
\begin{enumerate}[noitemsep]
    \item Move to $\delta(\tilde{a}_2)$ for pre-grasp.
    \item Move to $\tilde{a}_2$ and close gripper.
    \item Gradually interpolate like the motion planner toward $\tilde{a}_4$ for goal state.
\end{enumerate}

\subsection{\textbf{\textsc{PourWater} and \textsc{TransferWater}}}
\paragraph{Motion planner.} The motion planner of these two tasks consists of seven sub-task stages. We explain each stage as follows, beginning with the visual observation and ending with a consequent end effector pose:
\begin{enumerate}[noitemsep]
    \item Observe $o_1$. Move the end effector to the pre-grasp pose. Reach $a_1=\delta(a_2)$.
    \item Observe $o_2$. Move the end effector to reach the position for grasping. Reach $a_2$.
    \item Observe $o_3$. Close gripper. Reach $a_3$.
    \item Observe $o_4$. Lift the cup up to the target height for pouring. Reach $a_4$.
    \item Observe $o_5$. Translate the cup horizontally to the position for pouring. Reach $a_5$.
    \item Observe $o_6$. Gradually tilt the cup to pour water out until the goal condition is satisfied. We apply spherical linear interpolation on the action rotation to slow down and make fluids controllable. Excessive tilting would suddenly empty the water. Reach $a_6$.
    \item Observe $o_7$. Gradually rotate the cup back to an upright pose, similar to the previous stage. Reach $a_7$.
\end{enumerate}

\paragraph{Learning and evaluation.} We extract two observation-action pairs for two-phase learning: $o_1,a_2$ and $o_4,a_{5,6}$, where $a_{5,6}$ combines the translation of $a_5$ and rotation of $a_6$. During evaluation, the robot executes the action predictions $\tilde{a}_2,\tilde{a}_{5,6}$ similar to the motion planner. Note that we maintain the angular velocity to a constant, which ensures reproducibility of the amount of poured water, as well as distinctions among the rotations for different goal states.
\begin{enumerate}[noitemsep]
    \item Move to $\delta(\tilde{a_2})$ for pre-grasp.
    \item Move to $\tilde{a}_2$ and close gripper.
    \item Lift up from $\tilde{a}_2$ to the height of $\tilde{a}_{5,6}$.
    \item Translate horizontally to the position of $\tilde{a}_{5,6}$, with rotation unchanged.
    \item Gradually tilt the cup to $\tilde{a}_{5,6}$ by spherical linear interpolation (Slerp).
    \item Gradually rotate the cup back to the upright pose.
\end{enumerate}

\section{Data Collection}
\subsection{Human Annotations}
We collect human annotations for the positions of robots and interactive assets to ensure reasonable configurations in 3D scenes. For the rationality of such annotations, we ask human operators to teleoperate robots to complete tasks. We will first introduce the settings of robot teleoperation in \cref{sec:teleoperation} and then present the whole pipeline of data collection in \cref{sec:collect_pipeline}.

\subsubsection{Robot Teleoperation}\label{sec:teleoperation}
\begin{figure}[t!]
    \centering
    \includegraphics[width=\linewidth]{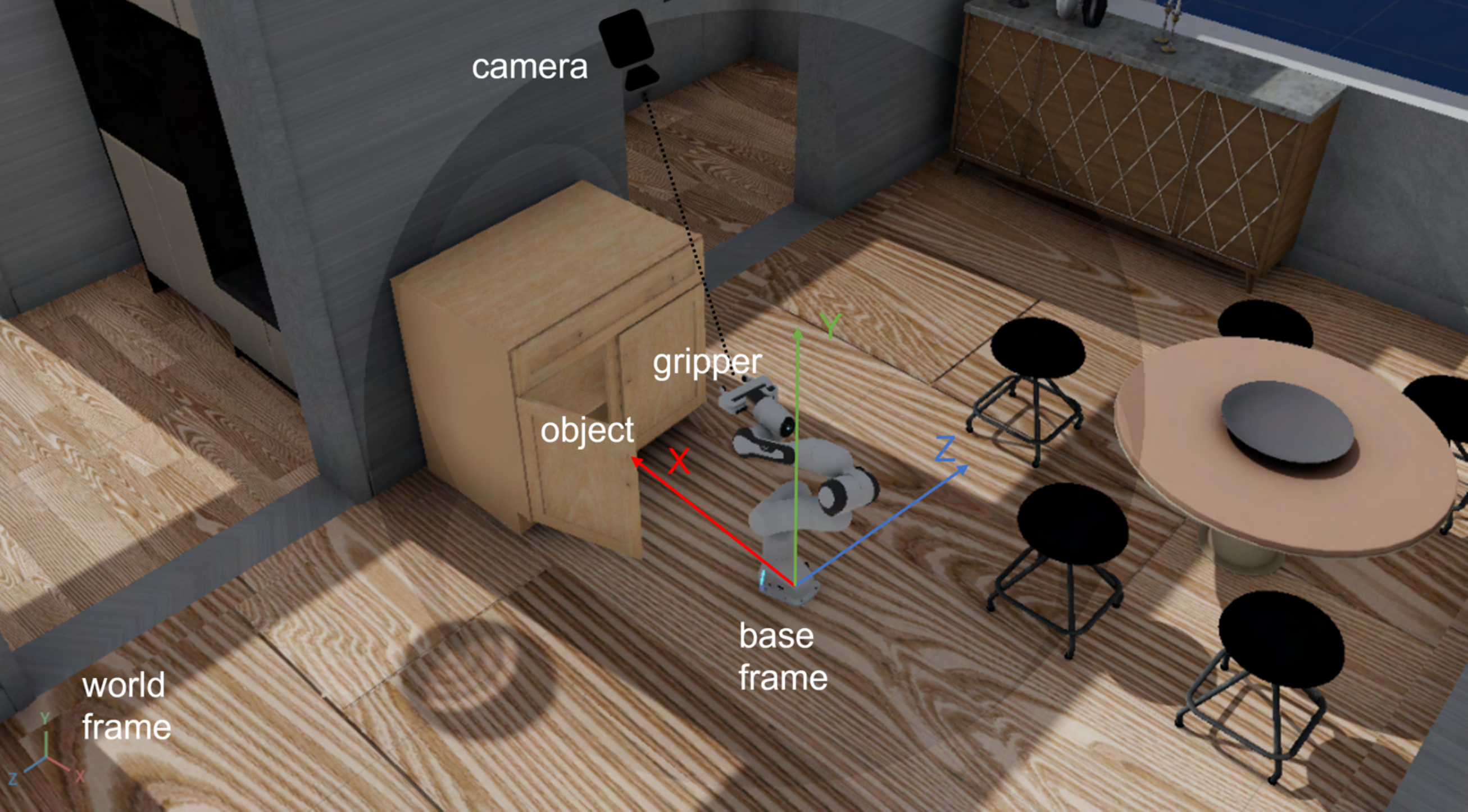}
    \caption{An illustration of the frame and camera for robot teleoperation.}
    \label{fig:frames}
\end{figure}

\paragraph{Frame definition.} Human operators use an Xbox controller to control the robot and the camera for data collection. The input to the controller is in the robot base frame, where the X axis originates from the robot's base and points to the object, and the Y axis is the upward pointing axis, as displayed in \cref{fig:frames}. This controller input is transformed into the world frame for robot motion at each timestep.

\begin{figure}[t!]
    \centering
    \includegraphics[width=\linewidth]{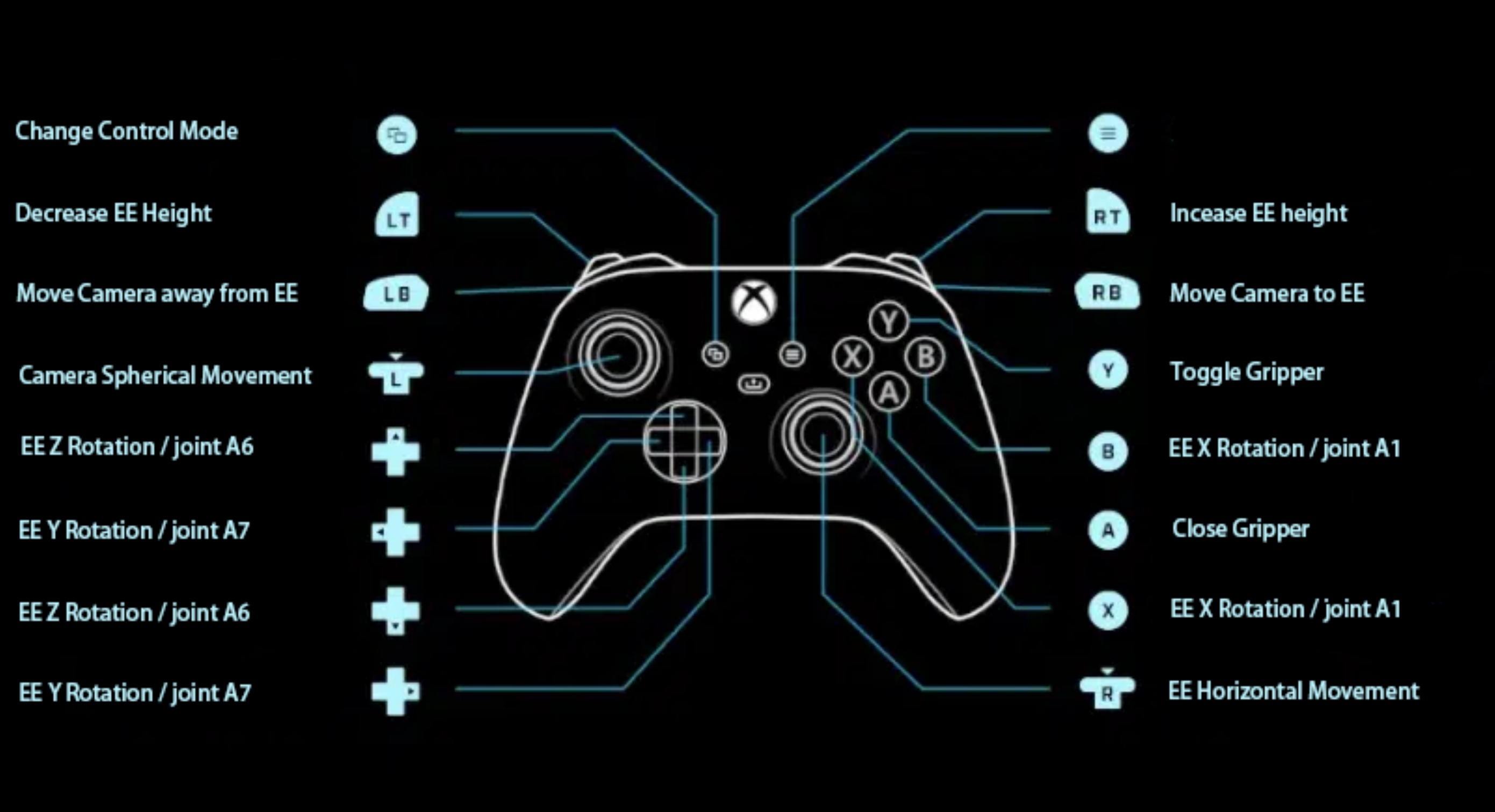}
    \caption{A schematic of the Xbox controller.}
    \label{fig:controller_mapping}
\end{figure}

\paragraph{Robot control.} Human operators can adjust the position and rotation of the robot end effector, as well as toggle the gripper. The Xbox layout is shown in \cref{fig:controller_mapping}. Specifically, the controller supports two control modes for rotation: joint position control and end effector rotation control. The joint position control mode allows the operator to directly change the positions of joints: A1 (shoulder joint), A6 (forearm joint), and A7 (wrist joint). The end effector rotation control mode allows the operator to rotate the end effector around the X, Y, Z axis of the robot base frame while maintaining its position, which is more general for rotation control. Operators can switch between these two modes during the data collection process.

\paragraph{Camera control.} Human operators can move their viewing camera freely in the 3D scenes to avoid occlusion. This is accomplished by a spherical camera control (\cref{fig:frames}), where the camera can move on a sphere centered at the robot end effector while keeping casting toward the end effector. Note that the radius of the sphere can also be adjusted for a clearer view.

\begin{figure}[t!]
    \centering
    \includegraphics[width=\linewidth]{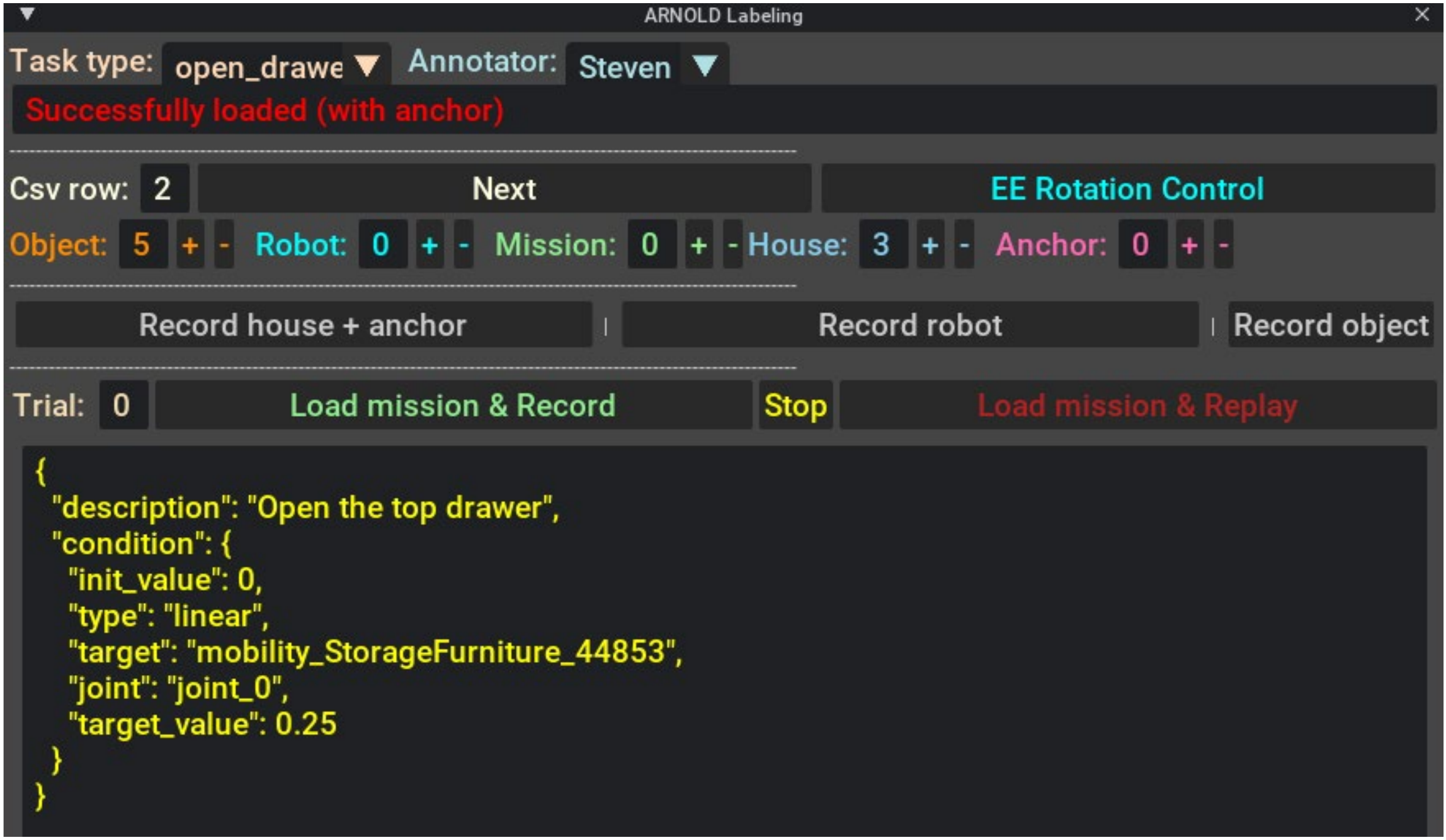}
    \caption{A toy example of the user interface (UI) for collecting human annotations.}
    \label{fig:UI}
\end{figure}

\subsubsection{Collection Pipeline}\label{sec:collect_pipeline}

\paragraph{Settings.} The collection of human annotations is conducted through a user interface (UI), which is implemented as an extension of Omniverse Isaac Sim, as shown in \cref{fig:UI}. For each task, we enumerate the compositions of objects, scenes, initial states and goal states. Each composition instance is called a mission. Human annotators are supposed to annotate each mission with two configurations of the relative positions between the robot and object. Each configuration is loaded for human control for two trials. Hence, each mission can produce up to four trajectories.

\paragraph{Procedures.} The annotator starts annotating a mission by placing the object group (robot and object) at an appropriate anchor position in the scene and clicking the ``Record house + anchor'' button. Then, the annotator is supposed to adjust the relative position between the robot and object before clicking the ``Record robot'' and ``Record object'' buttons. Clicking these two buttons amounts to recording a configuration. For each configuration, the annotator is supposed to control the robot to complete the task after clicking the ``Load mission \& Record'' button. Each click is regarded as a trial. Once the annotator succeeds, a message of ``Task Success'' will be displayed. Then the annotator can click the ``Stop'' button and thereby produce a human trajectory. The trajectory replay is supported by clicking the ``Load mission \& Replay'' button. Altogether, each mission is annotated with two configurations and two trials for each configuration. To proceed to the next mission, the annotator can click the ``Next'' button.

\begin{figure}[t!]
\centering
\hspace{-19pt}
    \includegraphics[width=0.9\linewidth]{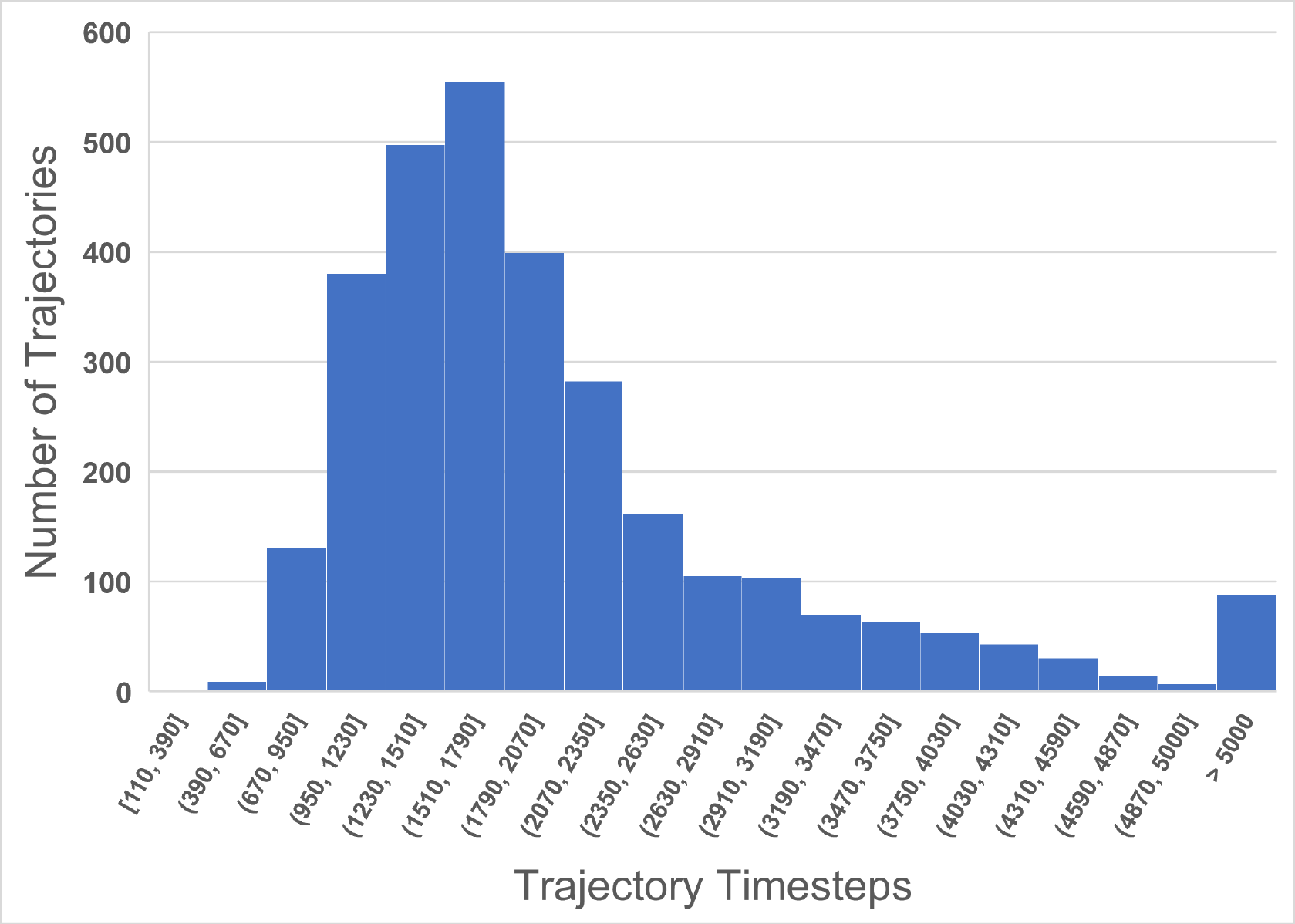}
    \caption{Distribution of human-annotated trajectory length (timesteps). Here $120$ timesteps amount to a second.}
    \label{fig:dataset_stats}
\end{figure}

\paragraph{Statistics.} We record the human trajectories with the aforementioned Xbox controller at $120$Hz and finally collect $2990$ trajectories. These trajectories consist of $6.2$M frames ($14.3$ hours) in total and $2073$ frames ($17.3$ seconds) on average. The minimum, median and maximum length are $110$ frames ($0.9$ seconds), $1752$ frames ($14.6$ seconds) and $11336$ frames ($94.5$ seconds), respectively. The distribution is shown in \cref{fig:dataset_stats}. Although these human trajectories are not directly used for training in \benchmark, the annotations of configurations and keypoints are referenced by motion planners for demonstration generation. And these human trajectories are of potential use when more powerful algorithms are available.

% We sample actions using an Xbox robot controller at $120$Hz resulting in 2,488 trajectories or around 4,400,000 frames of image/action pairs. The recorded trajectory length is about 10.2 hours, with a median trajectory length of 13 seconds, an average of 14.8 seconds, a maximum of 91.6 seconds, and a minimum of 3.2 seconds. Distribution shown in \cref{fig:dataset_stats}. 

\begin{table*}[t!]
\centering
\begin{tabular}{ll}
\toprule
Tasks         &  Examples of Delexicalized Templates \\
\midrule
\textsc{PickupObject} & \textit{Raise \textup{[value\_object] [value\_height]} above the ground} \\
\textsc{ReorientObject} & \textit{Reorient \textup{[value\_object] [value\_degree]} away from the up axis} \\
\textsc{OpenDrawer} & \textit{Open the \textup{[value\_position] [value\_object] [value\_percent]}} \\ 
\textsc{CloseDrawer} & \textit{Close the \textup{[value\_position] [value\_object] [value\_percent]}}  \\ 
\textsc{OpenCabinet} & \textit{Open the \textup{[value\_position] [value\_object] [value\_percent]}} \\
\textsc{CloseCabinet} & \textit{Close the \textup{[value\_position] [value\_object] [value\_percent]}} \\
\textsc{PourWater} & \textit{Pour \textup{[value\_percent]} water out of \textup{[value\_object]}} \\
\textsc{TransferWater} & \textit{Transfer \textup{[value\_percent]} water to \textup{[value\_object]}} \\
\bottomrule
% TapWater          &<v><\text{o}><s> & fill the cup entirely full \\
% \hline
\end{tabular}
\vspace{0.5em}
\caption{A few examples of delexicalized instruction templates for various tasks. During data generation, we first sample a delexicalized instruction template from the template pool. Then, we fill in the placeholders by sampling from candidate pools that consist of equivalent phrases.}
\label{tab:template2}
\end{table*}

\subsection{Data Augmentation}
For richer data variations, we apply augmentation to the robot positions based on collected human annotations. For example, the robot positions may be expanded by shifting $10$cm horizontally along four orthogonal directions. After that, we use the motion planner to check if the robot can execute the task successfully with new positions.

\subsection{Language instructions}
We sample a few instruction templates from our template pool and present them in \cref{tab:template2}. For generality and diversity, we construct these templates with placeholders, which can be lexicalized flexibly.
In these templates, ``[value\_object]'' holds for the actual object name, \eg, ``\textit{white bottle}''. The goal states are specified by ``[value\_height]'' (\eg, ``\textit{$20$cm}''), ``[value\_degree]'' (\eg, ``\textit{$90$ degrees}''), and ``[value\_percent]'' (\eg, ``\textit{$40\%$}''). In some tasks where multiple objects exist, the referential words are necessary and specified by ``[value\_position]'', which indicates the positional information of the target object, \eg, ``\textit{top left}''. The placeholders are lexicalized randomly by sampling from candidate pools which contain equivalent phrases, \eg, ``\textit{fifty percent}'', ``\textit{half}'', ``\textit{two quarters}''.

\subsection{Verification}
We replay the recorded keypoints of the generated demonstrations in the procedure of evaluation to verify their validity. After removing the failed demonstrations, we obtain a set of 10k demonstrations whose success is ensured achievable.

\section{Implementation Details}
\label{sec:implementation}

\subsection{Additional Evaluation of BC-Z}\label{sec:bc_z}
BC-Z \cite{jang2021bc} is a behavior-cloning model that pursues zero-shot task generalization with large-scale data. We follow the implementation of PerAct \cite{shridhar2022perceiver} to adapt the language-conditioned model (BC-Lang) to our settings. This model takes as input a single-view RGB-D image, with RGB and depth processed by two separate streams. We select the front view for the single-view visual input. The task instruction is processed by a CLIP text encoder \cite{radford2021learning} to extract a global semantic embedding. With the visual and language input, BC-Lang outputs directly regresses an end effector pose, whose translation and rotation are both continuous values (coordinates and quaternions). There are two variants of backbone: CNN and ViT. We run additional experiments with the two BC-Lang variants and report their performances on the \texttt{OpenDrawer} task (shown in \cref{tab:bc-z}). The results indicate BC-Z cannot handle the tasks in \benchmark.
\begin{table*}[t!]
    \centering
    \caption{The performances of two BC-Lang variants. We add the performance of PerAct for comparison. The results show that BC-Z, regressing end effector pose with a single-view image as visual input, is much less effective.}
    \vspace{0.5em}
    \small
    \resizebox{0.7\linewidth}{!}{
    \begin{tabular}{lrrrrrrrrrr}
        \multicolumn{1}{c}{} & \multicolumn{2}{c}{Test} & \multicolumn{2}{c}{Object} & \multicolumn{2}{c}{Scene} & \multicolumn{2}{c}{State} & \multicolumn{2}{c}{Any State} \\
        \midrule
        BC-Lang-CNN & 2.52 & \textcolor{NewGreen}{14.29} & 0.00 & \textcolor{NewGreen}{0.65} & 1.18 & \textcolor{NewGreen}{5.88} & 0.00 & \textcolor{NewGreen}{1.44} & 0.84 & \textcolor{NewGreen}{0.84} \\
        BC-Lang-ViT & 5.88 & \textcolor{NewGreen}{19.33} & 0.00 & \textcolor{NewGreen}{0.00} & 0.39 & \textcolor{NewGreen}{2.35} & 0.86 & \textcolor{NewGreen}{4.02} & 0.84 & \textcolor{NewGreen}{5.04} \\
        \midrule
        PerAct & 31.09 & \textcolor{NewGreen}{44.54} & 6.45 & \textcolor{NewGreen}{21.29} & 20.78 & \textcolor{NewGreen}{31.37} & 10.92 & \textcolor{NewGreen}{12.64} & 21.85 & \textcolor{NewGreen}{30.25}
    \end{tabular}
    }
    \label{tab:bc-z}
\end{table*}

\subsection{Workflow}
\paragraph{Data sampling.} We consider three aspects of balance when sampling data for training. (1) To ensure object-level balance, we categorize the demonstrations according to the manipulated object. A uniform sampling over categories is prior to sampling a demonstration within the category. (2) To ensure phase-level balance, we use a biased weight of 1:4 to sample a phase for training based on a selected demonstration. (3) To ensure task-level balance during multi-task training, a uniform sampling over tasks is prior to the task-specific sampling.

\paragraph{Validation.} We perform model selection by evaluating each checkpoint instead of tracking the loss on \textit{Val} set since we observe an inconsistency between these two metrics, as mentioned by \cite{shridhar2022cliport}. We select the best checkpoint for final evaluation.

\subsection{Ablation}
\paragraph{Without language.} We simply skip the operation of appending language embedding to the flattened voxel grid, making the latent representation contain only visual information.

\paragraph{State head.} As a naive implementation, we represent all object states through a normalized value in the $[0,1]$ range. For \textsc{PickupObject}, the state value is expressed by $\frac{h}{40}$ (initially $0$), where $h$ is the height in cm. For \textsc{ReorientObject}, the state value is expressed by $\frac{\alpha}{180}$ (initially $0.5$), where $\alpha$ is the angle (degrees) between object orientation and the upward axis. For the rest of the tasks, the state value is equal to the percentage. The state head predicts both the current state and the goal state, which are optimized via MSE loss. Although learning to regress state values brings moderate improvements, there is still quite a large space for better methods on state modeling.

\paragraph{Language encoder.} We adopt T5-base encoder \cite{raffel2020exploring} as the alternative to CLIP language encoder. To ensure a fair comparison, we pad the token sequence encoded by T5-base to $77$, the same as the token sequence of CLIP. Despite the sophisticated pre-training on large-scale text corpus, T5-base as a language encoder for language-grounded task may be limited by a lack of alignment with vision modality. Moreover, the scarce data in robotics tasks may induce T5-base to suboptimal performances.

\subsection{Sim2Real}\label{sec:sim2real}
\paragraph{Equipment and configurations.} To set up the real-robot experiment, we adopted the Franka Emika Panda robot arm, the RealSense D435 RGB-D camera, and a random common object in the object category. We placed the robot arm and object similar to the configurations in simulation. Notably, we only used the left camera view, and nothing was tuned particularly. We used the aruco marker to calibrate the D435 camera and Franka API to initialize the robot arm.

\paragraph{Inference.} We pre-processed the point cloud due to the imperfect depth camera, \eg, discarding noisy or too far away points. Next, we fetched the PerAct model trained in the simulator to perform inference for the next movement of the robot arm. We utilized Franka API to execute the actions and evaluate similarly to the metrics in \benchmark.

\paragraph{Results.} We summarize the real-world experiment as follows. (1) We experimented with opening/closing two different drawers and picking up five objects. We conducted 19, 8, and 31 trials in various configurations (\eg, object poses, camera poses, instructions \etc) for closing drawers, opening drawers, and picking up objects, respectively. And the corresponding numbers of success are 6, 0, and 4. (2) Throughout our experiments, we observe preliminary Sim2Real transfer capabilities, \ie, reasonable predictions for picking up objects and manipulating drawers. Nonetheless, the complex real-world environments, \eg, strict friction and sensory noise, still limit the performance of Sim2Real transfer to a considerable extent.

\end{document}